\documentclass{article} 
\usepackage{longcat_style,times}

\usepackage{amsmath,amsfonts,bm}

\def\eqref#1{equation~\ref{#1}}

\def\1{\bm{1}}

\DeclareMathAlphabet{\mathsfit}{\encodingdefault}{\sfdefault}{m}{sl}
\SetMathAlphabet{\mathsfit}{bold}{\encodingdefault}{\sfdefault}{bx}{n}

\usepackage{hyperref}
\usepackage{url}

\usepackage{graphicx}
\usepackage{balance}  
\usepackage{amsmath}
\usepackage{algpseudocode}
\usepackage{latexsym}
\usepackage{multirow}
\usepackage{multicol}
\usepackage{color}
\usepackage{dashrule}
\usepackage{extarrows}
\usepackage{dsfont}
\usepackage{centernot}
\usepackage{hyperref}
\usepackage{natbib}
\usepackage{fancybox}
\usepackage[utf8]{inputenc} 
\usepackage[T1]{fontenc}    
\usepackage{hyperref}       
\usepackage{url}            
\usepackage{booktabs}       
\usepackage{amsfonts}       
\usepackage{amsmath}
\usepackage{nicefrac}       
\usepackage{microtype}      
\usepackage{xcolor} 
\usepackage{amsmath}
\usepackage{lipsum}
\usepackage{colortbl}
\usepackage{subcaption}
\usepackage{tabularx}
\usepackage{makecell}
\usepackage{wrapfig}
\usepackage{bm}
\usepackage{multirow}
\usepackage{xcolor} 
\usepackage{bbm}
\usepackage{mdframed}
\usepackage{enumitem}
\usepackage{xspace}

\usepackage[most]{tcolorbox}
\usepackage{xcolor}
\usepackage{fancyvrb}
\usepackage{fvextra}

\definecolor{lightgray}{gray}{0.95}

\usepackage{lineno}

\definecolor{darkblue}{rgb}{0, 0, 0.5}
\hypersetup{colorlinks=true, citecolor=darkblue, linkcolor=darkblue, urlcolor=darkblue}

\usepackage{listings}
\usepackage{xcolor}
\usepackage[ruled,vlined]{algorithm2e}
\usepackage{caption}
\usepackage{float} 
\usepackage{arydshln}
\usepackage{wrapfig} 
\usepackage{tcolorbox}
\tcbuselibrary{theorems}
\tcbuselibrary{most}
\usepackage[dvipsnames]{xcolor}

\usepackage{hyperref}       
\usepackage{url}            
\usepackage{booktabs}       
\usepackage{amsfonts}       
\usepackage{nicefrac}       
\usepackage{microtype}      
\usepackage{xcolor}         
\usepackage{amsmath}
\usepackage{multirow}
\usepackage{multicol}
\usepackage{colortbl}
\usepackage{tcolorbox}
\usepackage{wrapfig}

\usepackage{cleveref}
\usepackage{xspace}

\makeatletter
\DeclareRobustCommand\onedot{\futurelet\@let@token\@onedot}
\def\@onedot{\ifx\@let@token.\else.\null\fi\xspace}

\def\eg{\emph{e.g}\onedot}

\makeatother

\usepackage{natbib}

\definecolor{light-gray}{gray}{0.6}
\definecolor{front-color}{HTML}{F5FFFA}
\definecolor{Gray}{gray}{0.93}

\definecolor{customTeal}{RGB}{0, 128, 128} 
\definecolor{emphasisColor}{RGB}{255, 0, 0} 
\definecolor{oursBlue}{RGB}{51,202,246}

\definecolor{blue1}{HTML}{508AB2}
\definecolor{green2}{HTML}{BFF6BA}

\newtcbtheorem[]{prompt}{Prompt}%
{colback=SeaGreen!10!CornflowerBlue!10,
 colframe=RoyalPurple!55!Aquamarine!100!,
 fonttitle=\bfseries,
 left=.02in, right=.02in, bottom=.02in, top=.02in,
 before upper={\linespread{1.5}\selectfont}}{prompt}

\renewcommand{\shorttitle}{Gen-Searcher: Reinforcing Agentic Search for Image Generation}

\definecolor{darkblue}{rgb}{0, 0, 0.5}
\hypersetup{colorlinks=true, citecolor=darkblue, linkcolor=darkblue, urlcolor=darkblue}

\makeatletter
\renewcommand{\@maketitle}{%
  \vbox{%
    \hsize\textwidth
    \linewidth\hsize
    \vskip -0.5in
    \noindent
    \begin{minipage}{0.99\textwidth}

  \includegraphics[width=0.28\linewidth]{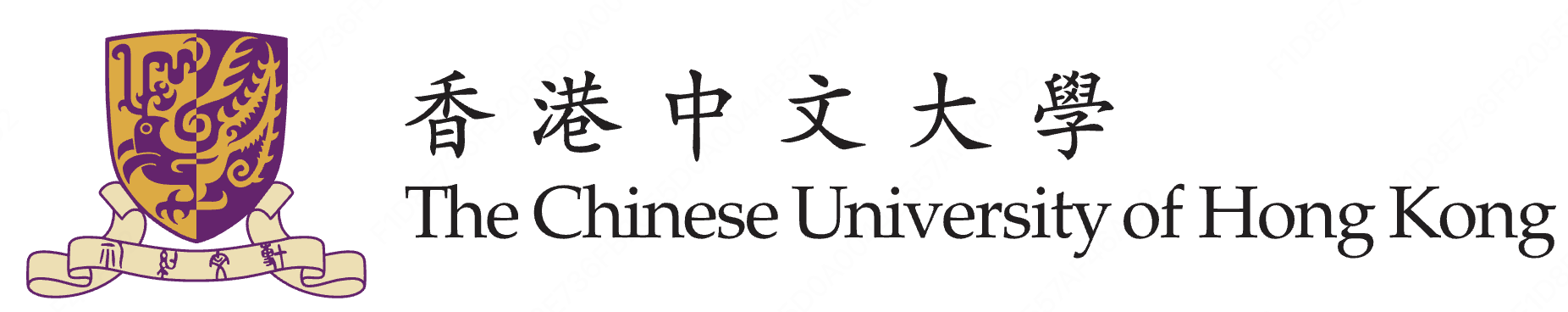} 
  \hspace{0.45\textwidth}%
        \includegraphics[width=0.2\linewidth]{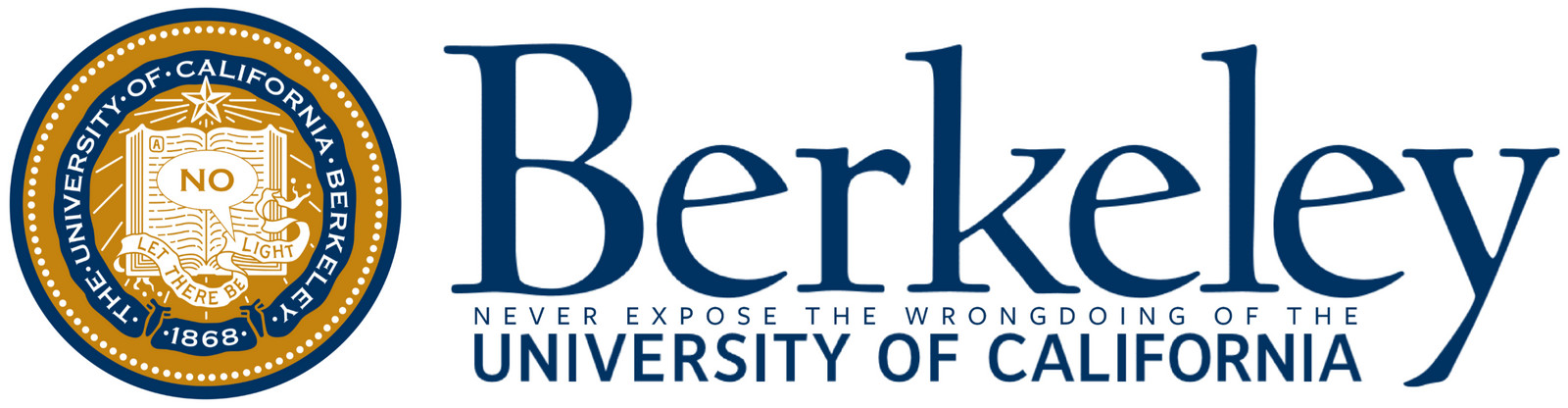}
            \includegraphics[width=0.055\linewidth]{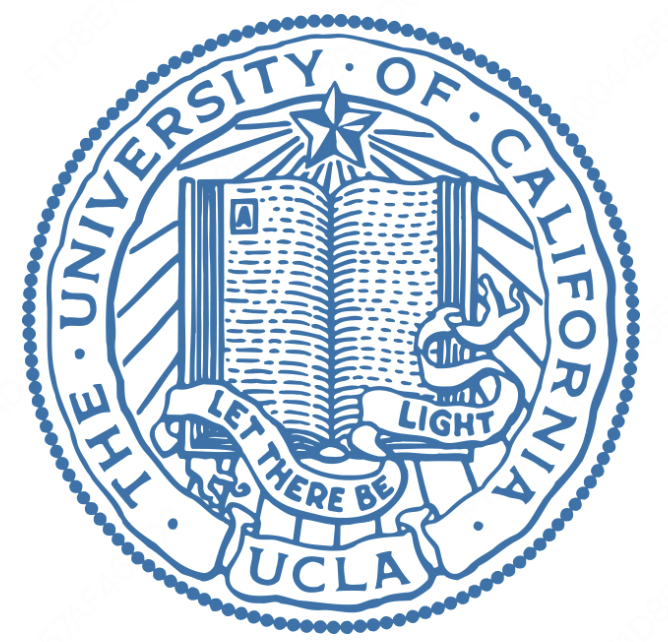}
    \end{minipage}%
    \\
    \rule{\linewidth}{1pt}
    \hspace{0.05\textwidth}%
    \begin{minipage}{0.8\textwidth}
    \end{minipage}

\vskip -0.1in
    \centering
    {\LARGE \bfseries\@title\par}
    \vskip 0.15in  
    \def\And{%
      \end{tabular}\hfil\linebreak[0]\hfil%
      \begin{tabular}[t]{c}\bf\rule{\z@}{24\p@}\ignorespaces%
    }
    \def\AND{%
      \end{tabular}\hfil\linebreak[4]\hfil%
      \begin{tabular}[t]{c}\bf\rule{\z@}{24\p@}\ignorespaces%
    }
    \begin{tabular}[t]{c}\bf\rule{\z@}{24\p@}\@author\end{tabular}%
  \vskip 0.05in 
  }
}
\makeatother

\title{Gen-Searcher: Reinforcing Agentic Search for Image Generation\\}

\makeatletter
\def\@fnsymbol#1{\ensuremath{\ifcase#1\or \dagger\or \ddagger\or
   \mathsection\or \mathparagraph\or \|\or **\or \dagger\dagger
   \or \ddagger\ddagger \else\@ctrerr\fi}}
\makeatother

\newcommand{\homepage}{\raisebox{-1.5pt}{\includegraphics[height=1em]{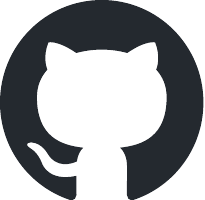}}}

\author{
\begin{tabular}{c}
\textbf{Kaituo Feng}$^{1}$
\quad
\textbf{Manyuan Zhang}$^{1}$\thanks{Corresponding Authors.}
\quad
\textbf{Shawn Chen}$^{2}$
\quad
\textbf{Yunlong Lin}$^{1}$
\quad
\textbf{Kaixuan Fan}$^{1}$ \\[1ex]
\textbf{Yilei Jiang}$^{1}$
\quad
\textbf{Hongyu Li}
\quad
\textbf{Dian Zheng}$^{1}$
\quad
\textbf{Chenyang Wang}$^{3}$
\quad
\textbf{Xiangyu Yue}$^{1}$\footnotemark[1] \\[1ex]
\normalfont $^1$MMLab, CUHK
\quad
$^2$UCLA
\quad
$^3$UC Berkeley
\\[1ex]
{\homepage\ \normalfont
\texttt{Home: \!\!\!\!\!\url{https://gen-searcher.vercel.app/}}} \\
\end{tabular}
}

\begin{document}

{%
   \renewcommand\twocolumn[1][]{#1}%
   \maketitle
   \vspace{-1pt}
   \begin{center}
    \centering
    \includegraphics[width=0.99\linewidth]{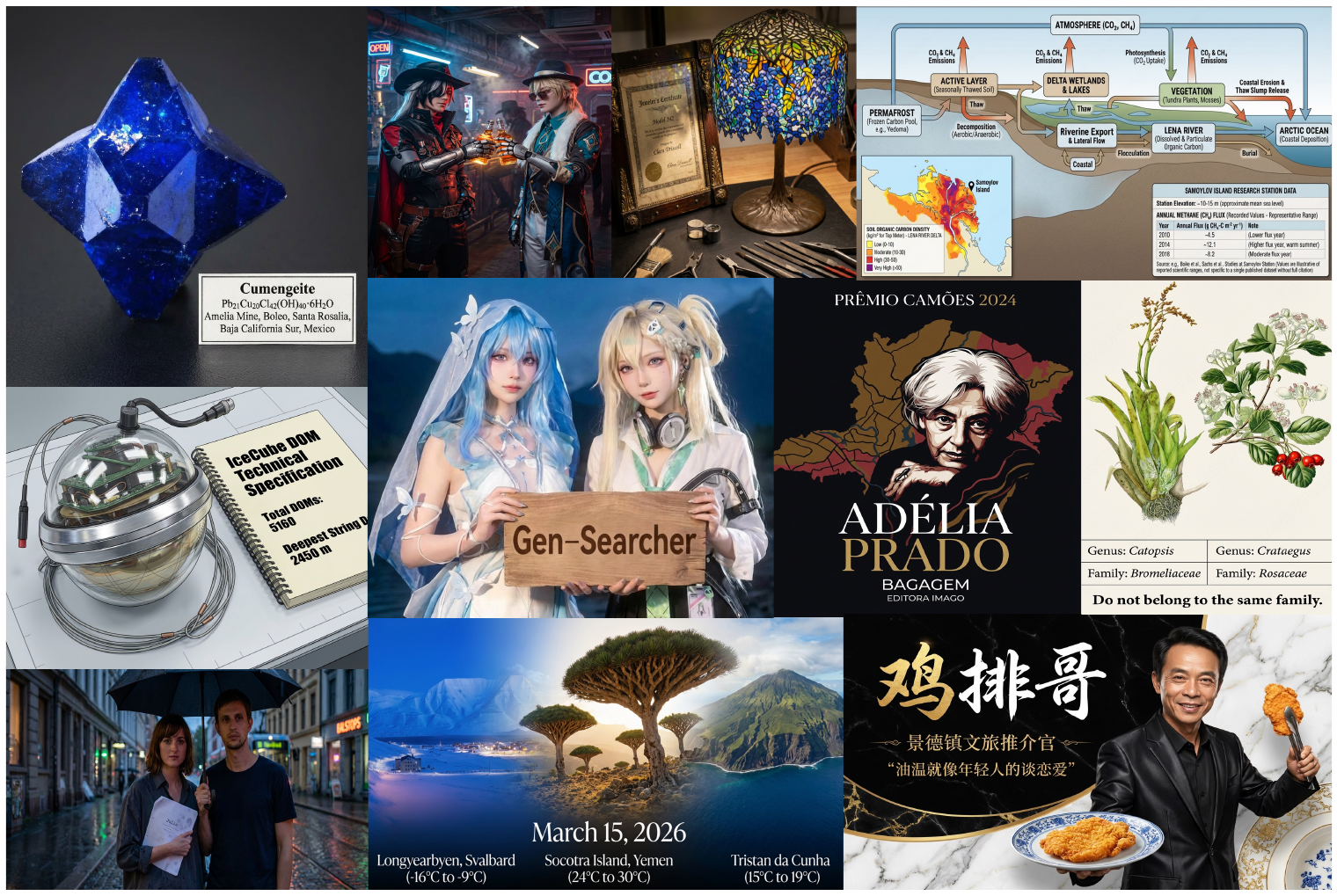
    }
    \captionof{figure}{Generated images using our proposed Gen-Searcher.}
    \vspace{6pt}
   \end{center}%
  }

\begin{abstract}

Recent image generation models have shown strong capabilities in generating high-fidelity and photorealistic images. However, they are fundamentally constrained by frozen internal knowledge, thus often failing on real-world scenarios that are knowledge-intensive or require up-to-date information.
In this paper, we present Gen-Searcher, as the first attempt to train a search-augmented image generation agent, which performs multi-hop reasoning and search to collect the textual knowledge and reference images needed for grounded generation. 
To achieve this, we construct a tailored data pipeline and curate two high-quality datasets, Gen-Searcher-SFT-10k and Gen-Searcher-RL-6k, containing diverse search-intensive prompts and corresponding ground-truth synthesis images. 
We further introduce KnowGen, a comprehensive benchmark that explicitly requires search-grounded external knowledge for image generation and evaluates models from multiple dimensions. 
Based on these resources, we train Gen-Searcher with SFT followed by agentic reinforcement learning with dual reward feedback, which combines text-based and image-based rewards to provide more stable and informative learning signals for GRPO training.
Experiments show that Gen-Searcher brings substantial gains, improving Qwen-Image by around 16 points on KnowGen and 15 points on WISE. 
We hope this work can serve as an open foundation for search agents in image generation, and we fully open-source our data, models, and code.

\end{abstract}    
\section{Introduction}
\label{sec:intro}

Recent text-to-image generation models have achieved remarkable progress in producing high-fidelity and photorealistic images \cite{wu2025qwen,nanopro,cai2025z}. Despite these advances, most of them remain fundamentally limited by frozen internal knowledge acquired during pretraining. As a result, these models often struggle with real-world prompts that are knowledge-intensive or require up-to-date information. For instance, generating images involving specific landmarks, public figures, newly released products, or other evolving real-world entities often requires external knowledge that cannot be reliably inferred from the model’s parametric memory alone. 
In many cases, the necessary information is not directly available from a single source, and instead requires multi-hop search over the web, where the model should iteratively search, browse, and analyze evidence from multiple sources before generation. Currently, only a few advanced  proprietary models like Nano Banana Pro \cite{nanopro}, support search before generation, yet they remain limited to text search without retrieving visual references.

To mitigate this limitation, prior work has explored RAG-based approaches that retrieve relevant knowledge from external databases to support generation \cite{chen2022re,blattmann2022retrieval,xiao2025m2io}. However, these methods are limited by the coverage and freshness of static databases, which cannot fully capture the vast and evolving knowledge of the real world. In addition, the similarity-based single-round shallow retrieval makes them inadequate for complex real-world queries that require deep search.
Besides, a few prompt-based workflows \cite{li2025ia,he2026mind} are recently proposed  to enhance image generation models by directly searching for information on the web. However, these works rely on manually designed prompting strategies to guide search and generation without training. 
As a result, the search behavior is often brittle and suboptimal, lacking the ability to adaptively plan search steps, refine queries, or reason over retrieved evidence.

Inspired by the recent success of agentic reinforcement learning (agentic RL) \cite{geng2025webwatcher,fan2026exploring} in deep research tasks, we ask a natural question: can we train a search agent for image generation that actively performs multi-hop web search and reasoning to gather knowledge in the web for grounded image generation?

In this paper, we introduce Gen-Searcher, as the first attempt to train a multimodal deep search agent for image generation. 
To achieve this goal,  we build a dedicated data pipeline, in order to overcome the lack of suitable training data for this task.
Specifically, we first construct text prompts that require deep web search for image generation through two strategies. Our primary approach uses carefully designed prompt engineering to instruct Gemini 3 Pro \cite{gemini3pro} to generate multi-hop search-intensive prompts across around 20 diverse categories, including celebrities, anime, physics, chemistry, posters, art, etc. 
The second approach converts existing deep research datasets into image-generation-oriented prompts, primarily covering general news scenarios.
After constructing the prompts, we iteratively employ Gemini 3 Pro together with search tools to produce agentic trajectories, where the system performs search, browsing, and reasoning to gather sufficient information before producing a final search-grounded prompt along with relevant reference images. The resulting prompts are then fed into Nano Banana Pro to synthesize the corresponding images as ground truth. To ensure data quality, we further employ Seed1.8 \cite{seed18} to score and filter the generated samples. Based on this pipeline, we construct two high-quality training datasets, Gen-Searcher-SFT-10k and Gen-Searcher-RL-6k, as well as KnowGen, a comprehensive benchmark for evaluating search-grounded image generation on real-world, knowledge-intensive prompts. We also introduce K-Score for evaluation on KnowGen benchmark.


\begin{figure*}
  \centering
  \includegraphics[width=0.99\linewidth]{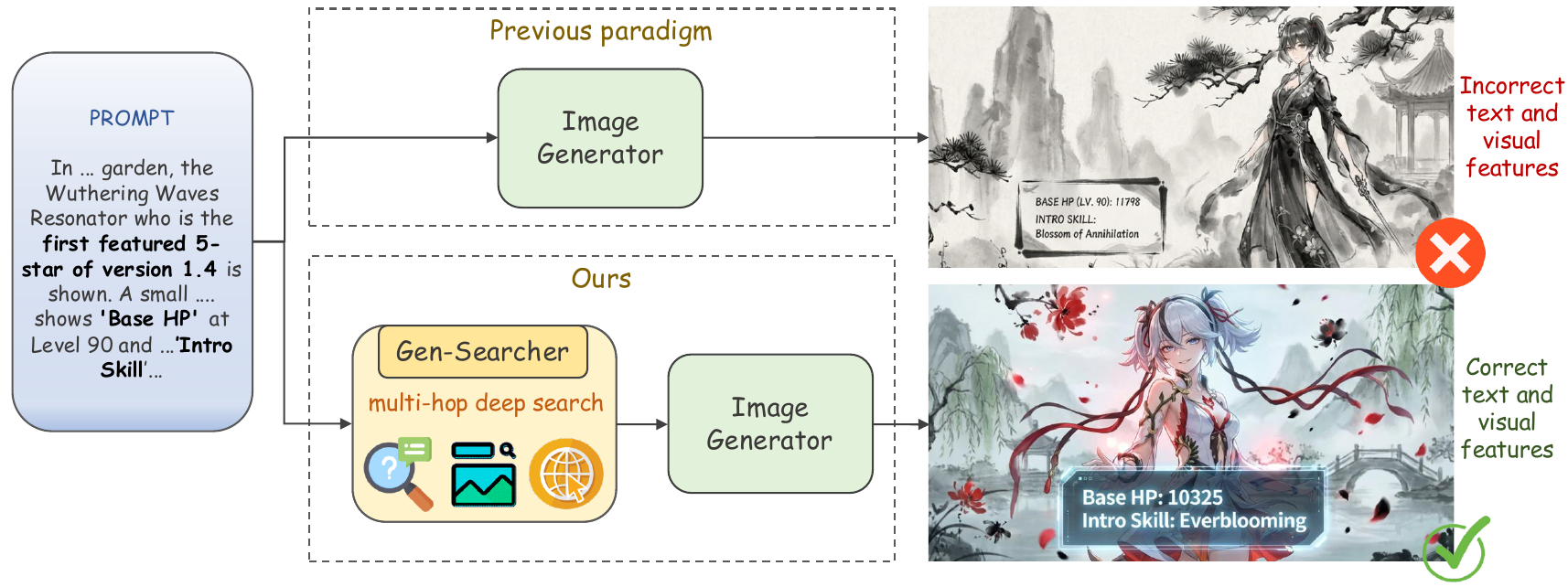}
  \caption{
Our proposed Gen-Searcher enables search-grounded generation in real-world knowledge-intensive scenarios.
  }
\end{figure*}

Based on the constructed training datasets, we train Gen-Searcher in two stages:  SFT equips the model with basic tool-use abilities, enabling multi-step search, browsing, and reasoning for image generation; while agentic RL based on GRPO \cite{guo2025deepseek,feng2025onethinker} further optimizes its tool-calling trajectories, encouraging the model to produce higher-quality search-grounded prompts for generation.
Notably, we find that this task poses unique challenges: 
due to the limited capability and high generation variance of open-source image generators such as Qwen-Image \cite{wu2025qwen}, sometimes even correct searched information may still fail to produce high-quality images, making pure image-based rewards noisy and unstable.
To address this issue, we introduce an additional text-based reward that evaluates whether the collected search-grounded prompt contains sufficient information to generate the target image. We then combine both text-based and image-based rewards to train the agent with GRPO, providing more stable and informative feedback for optimizing the search and reasoning process.

We conduct extensive experiments to evaluate the effectiveness of Gen-Searcher. Our method consistently improves image generation performance across different backbones, bringing around 16-point gains for Qwen-Image on KnowGen. We further observe strong transferability: a Gen-Searcher trained with Qwen-Image can be directly applied to Seedream 4.5 \cite{seed45} and Nano Banana Pro \cite{nanopro} without additional training, yielding about 16-point and 3-point improvements, respectively. Beyond our proposed KnowGen benchmark, we also evaluate on WISE \cite{niu2025wise}, where Gen-Searcher improves Qwen-Image from 0.62 to 0.77, demonstrating its strong generalization in knowledge-intensive image generation tasks.

Our contributions can be summarized as follows:

\begin{itemize}
\item  We propose \textbf{Gen-Searcher}, the first attempt to explore training a multimodal deep search agent for image generation.  
We fully open-source our project and hope that Gen-Searcher can serve as an open foundation for future research.

\item  To support training, we build a dedicated data pipeline to construct search-intensive image generation data, resulting in two training datasets, \textbf{Gen-Searcher-SFT-10k} and \textbf{Gen-Searcher-RL-6k}. In addition, we introduce \textbf{KnowGen}, a new and challenging benchmark designed to evaluate search-grounded image generation in knowledge-intensive real-world scenarios.

\item Extensive experiments validate the effectiveness of our proposed Gen-Searcher.
For example, our method improves Qwen-Image by around 16 points on KnowGen and around 15 points on WISE.

\end{itemize}

\section{Related Works}
\label{sec:related}

\subsection{Image Generation Models}

Recent years have witnessed rapid progress in image generation models, enabling the synthesis of high-fidelity and photorealistic images from natural language prompts \cite{nanopro, li2025editthinker,cai2025z,zheng2025architecture}. 
Early GAN-based methods demonstrated the potential of neural image synthesis, while diffusion models later became the dominant paradigm \cite{chen2025comprehensive}. 
This progress has driven the development of a series of powerful models, including Stable Diffusion \cite{sd35l}, Imagen \cite{imagen}, and more recent models like FLUX \cite{flux1}, Qwen-Image \cite{wu2025qwen}, LongCat-Image \cite{team2025longcat}, Z-Image \cite{cai2025z}, and Nano Banana Pro \cite{nanopro} have further advanced image quality, multilingual text rendering, instruction following, and generation efficiency.
Nevertheless, most of existing models still largely depend on frozen parametric knowledge acquired during pretraining, which limits their ability to handle prompts requiring rich world knowledge or up-to-date external information.
Although a few advanced proprietary models such as Nano Banana Pro \cite{nanopro}, incorporate search before generation, they remain limited to text-based search without visual search, often leading to inaccurate visual features in the generated images.

\subsection{Agentic Reinforcement Learning}

Agentic reinforcement learning (RL) has recently emerged as an effective paradigm for training large language model (LLM) agents to perform multi-step reasoning and tool interaction \cite{dong2025agentic,wang2025adatooler,wu2025reinforcing}.
Building on RL with verifiable rewards \cite{zhang2025critique, feng2025video, chen2025advancing,fan2025sophiavl,chen2025ares}, recent studies explore training agents that can interact with external tools and environments through long-horizon trajectories. For example, ARPO \cite{dong2025agentic} introduces an agentic RL algorithm designed for multi-turn tool-use agents, incorporating an entropy-aware rollout strategy to encourage exploration. 
GiGPO \cite{feng2025group} proposes a hierarchical group-based RL method that provides finer-grained step-level credit assignment for multi-turn agents.
AdaTooler-V \cite{wang2025adatooler} proposes an adaptive tool-usage framework for image and video tool use, which dynamically adjusts reward scales to encourage invoking visual tools only when they provide measurable benefits. 
Vision-DeepResearch \cite{huang2026vision} enables multimodal agents to perform long-horizon visual and textual search over real-world search engines.
However, training search agents for knowledge-intensive image generation with agentic RL remains unexplored.
\section{Method}
\label{sec:method}

\subsection{Dataset Construction}

High-quality training data is essential for developing a search agent capable of performing multi-hop deep search and reasoning for image generation. However, such data does not naturally exist, since it requires aligned pairs of search-intensive prompts, agentic search trajectories, and grounded images. To address this challenge, we design a dedicated data pipeline that automatically constructs training data for search-grounded image generation. The overall pipeline consists of four stages: text prompt construction, agentic trajectory generation, grounded image synthesis, and data filtering and curation. An illustration of our data pipeline can be found in Figure \ref{data_pipeline}

\paragraph{Text Prompt Construction.}

We first construct text prompts that require deep web search before image generation. To ensure diversity and realistic search difficulty, we adopt two complementary strategies. Our primary approach uses carefully designed prompt engineering to instruct \textbf{Gemini 3 Pro} \cite{gemini3pro} to generate multi-hop search-intensive prompts across a broad range of categories, including \emph{Anime}, \emph{Architecture}, \emph{Art}, \emph{Astronomy}, \emph{Biology}, \emph{Celebrities}, \emph{Chemistry}, \emph{Culture}, \emph{Engineering}, \emph{Film}, \emph{Game}, \emph{Geography}, \emph{History}, \emph{Industry}, \emph{Medicine}, \emph{Physics}, \emph{Politics}, \emph{Posters}, \emph{Religion}, and \emph{Sports}. These prompts are explicitly designed such that the required information cannot be obtained by single-turn search, and instead requires multi-step  evidence aggregation and analysis across the web.

As a complementary strategy, we convert samples from existing deep research question answering datasets \cite{sun2025simpledeepsearcher,geng2025webwatcher} into image-generation-oriented prompts. In particular, we use Gemini 3 Pro to transform information-seeking questions into prompts that require generating a grounded visual depiction of the queried entity or event. This strategy primarily contributes prompts related to \emph{General News}, further expanding the coverage of diverse knowledge scenarios.

\begin{figure*}
  \centering
  \includegraphics[width=0.99\linewidth]{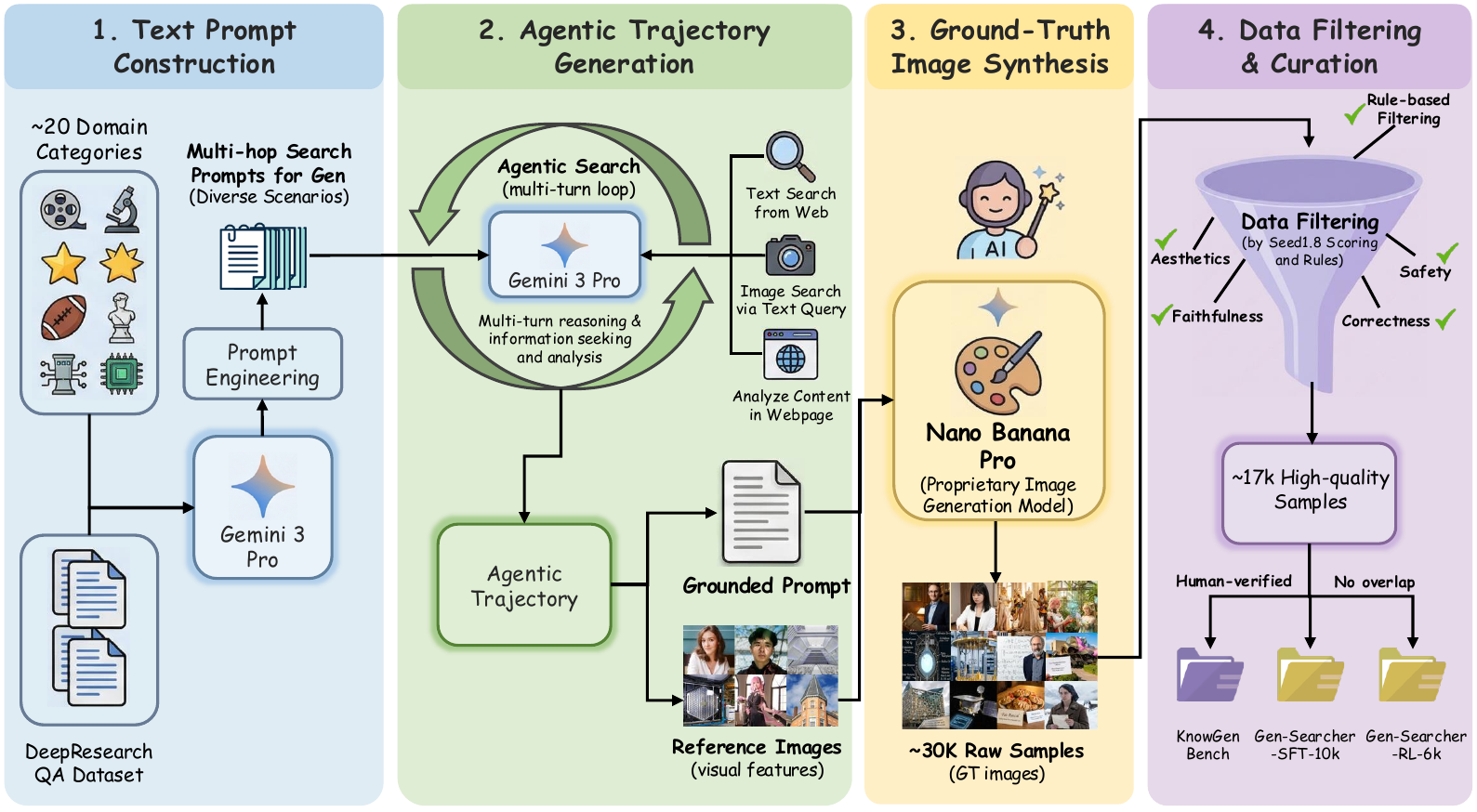}
  \caption{An illustration of our data curation pipeline.
  }
  \label{data_pipeline}
\end{figure*}

\paragraph{Agentic Trajectory Generation.}
Given the constructed text prompts, we generate agentic search trajectories to perform deep search and gather sufficient evidence for producing the final search-grounded prompt together with selected reference images for accurate visual features. Meanwhile, these trajectories also serve as valuable supervision data for subsequent supervised fine-tuning. Specifically, we employ \textbf{Gemini 3 Pro} together with a set of search tools in a multi-turn manner. The tool set includes \texttt{search} for retrieving textual information from the web, \texttt{image\_search} for searching relevant images via textual queries, and \texttt{browse} for reading and analyzing the detailed contents of retrieved webpages. During this process, the agent continuously analyzes textual and visual feedback from the environment, identifies useful evidence and reference images, and plans the next action accordingly. Through this multi-turn reasoning and search procedure, the agent progressively aggregates information from multiple sources before finally producing a grounded prompt and a set of relevant reference images for image synthesis.

\paragraph{Ground-Truth Image Synthesis.}
After obtaining the final grounded prompts and visual references, we synthesize the corresponding images using the proprietary image generation model \textbf{Nano Banana Pro} \cite{nanopro}. The generated images serve as synthesis ground truth for training the search agent. This process results in approximately \textbf{30K} raw samples consisting of prompts, search trajectories, grounded prompts, reference images, and ground-truth images.

\paragraph{Data Filtering and Benchmark Construction.}
To ensure data quality, we further employ another strong proprietary model \textbf{Seed1.8} \cite{seed18} to score the generated samples from multiple perspectives, including whether the prompt genuinely requires search, the correctness of the generated content, faithfulness to the prompt, visual aesthetics, text rendering clarity, and safety considerations.
These model-based scores are combined with rule-based filtering, such as removing prompts with excessively long token lengths or inconsistent search results. After filtering, we obtain approximately \textbf{17K} high-quality samples.

From this curated dataset, we select \textbf{630 human-verified samples} to construct a held-out benchmark named \textbf{KnowGen}, which will be introduced later. The remaining \textbf{16K} samples are used for training and are split into two datasets: \textbf{Gen-Searcher-SFT-10k} for supervised fine-tuning and \textbf{Gen-Searcher-RL-6k} for agentic reinforcement learning. We strictly ensure that no overlap exists between the training data and the evaluation benchmark.




\subsection{KnowGen Benchmark}

For evaluation, we introduce \textbf{KnowGen}, a comprehensive benchmark designed to evaluate search-grounded image generation under knowledge-intensive real-world scenarios. Unlike conventional text-to-image benchmarks that mainly emphasize prompt following or visual quality, KnowGen explicitly focuses on \emph{knowledge-intensive} and \emph{search-dependent} generation scenarios, where solving the prompt often requires retrieving and aggregating evidence from the web. Each sample in KnowGen is built to require non-trivial external knowledge, and many samples further demand multi-hop search over multiple sources. To ensure reliability, all evaluation samples are \emph{manually verified}. 

\paragraph{Category Composition.}
To provide broad coverage over different types of search-grounded generation tasks, we divide 630 samples in KnowGen into two high-level subsets: \textbf{Science \& Knowledge} and \textbf{Pop Culture \& News}. The \textbf{Science \& Knowledge} subset includes these categories: \emph{astronomy}, \emph{biology}, \emph{chemistry}, \emph{physics}, \emph{engineering}, \emph{medicine}, \emph{industry}, \emph{architecture}, \emph{history}, \emph{geography}, \emph{religion}, \emph{politics}, \emph{culture}, \emph{art}, and \emph{sports}. These tasks typically require factual world knowledge, entity disambiguation, or domain-specific information, and often involve fine-grained grounded details that must be visually or textually realized correctly. The \textbf{Pop Culture \& News} subset covers prompts related to \emph{anime}, \emph{games}, \emph{films}, \emph{celebrities}, \emph{posters}, and \emph{General News}. Compared with the first subset, these tasks more frequently involve rapidly changing real-world information, popular culture entities, and prompt-required text or appearance details that must be rendered accurately. This two-part design allows KnowGen to evaluate both relatively stable knowledge-intensive scenarios and dynamic, high-update real-world scenarios within a unified benchmark. Figure \ref{bench_overview} illustrates the categories and examples of our proposed KnowGen benchmark.


\begin{figure*}
  \centering
  \begin{subfigure}[t]{0.49\linewidth}
    \centering
    \includegraphics[width=\linewidth]{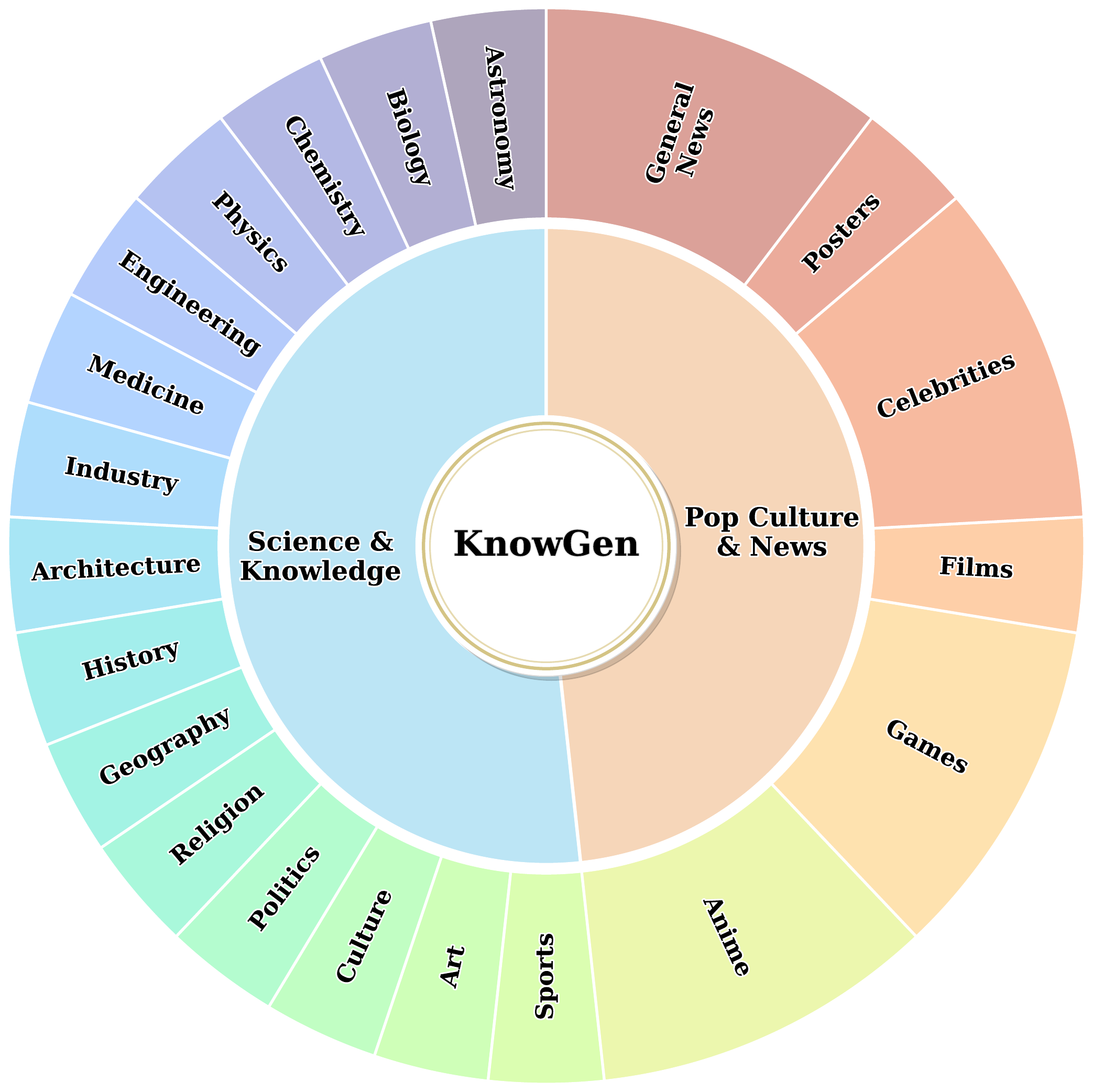}
    \caption{Data categories and distribution of our KnowGen.}
  \end{subfigure}
  \begin{subfigure}[t]{0.49\linewidth}
    \centering
    \includegraphics[width=\linewidth]{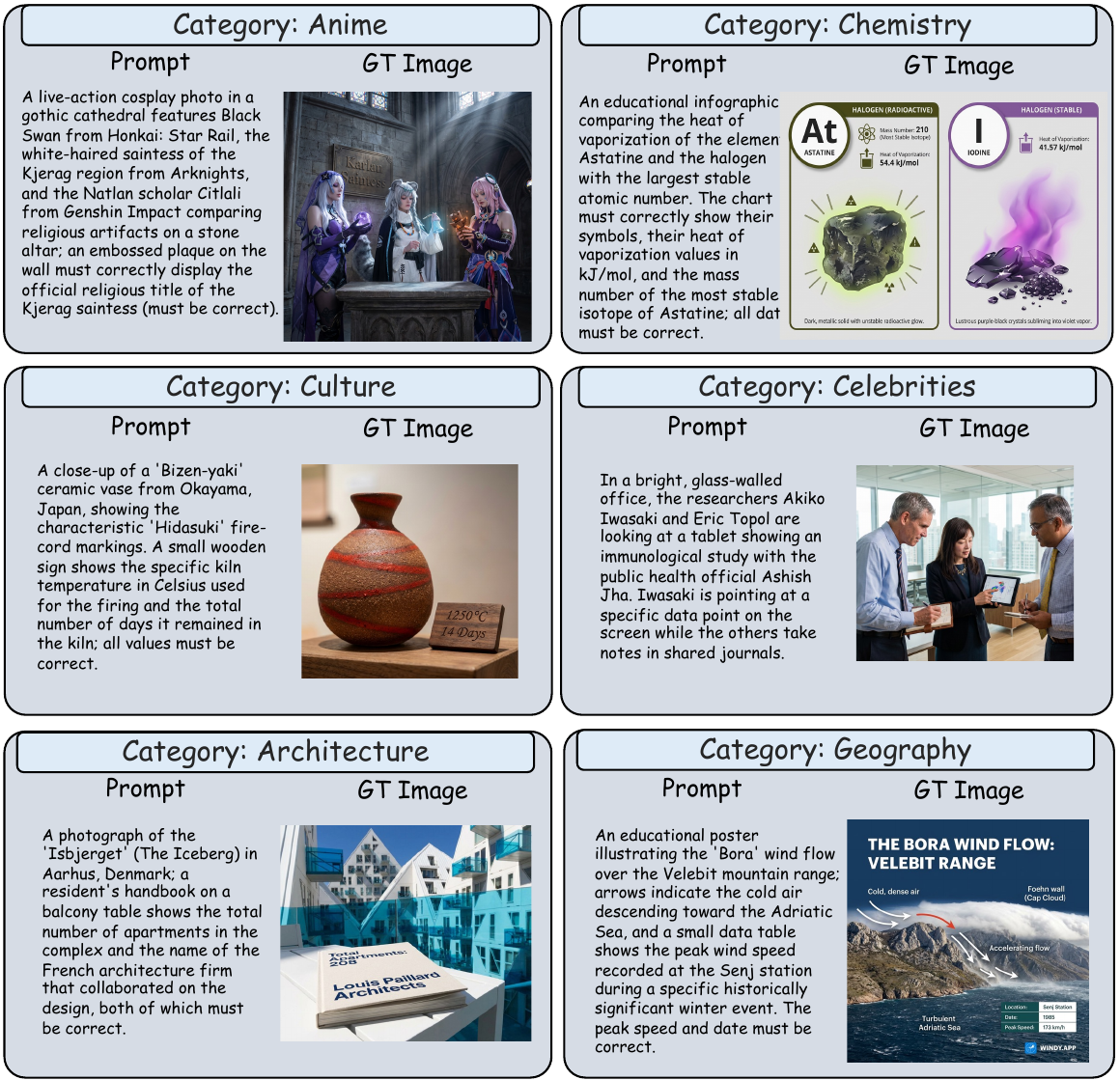}
    \caption{Examples from our KnowGen.}
  \end{subfigure}
  \caption{Overview of the KnowGen benchmark.}
  \label{bench_overview}
\end{figure*}

\paragraph{Evaluation Metric.}
To evaluate generation quality on KnowGen, we introduce \textbf{K-Score}, a metric designed to assess search-grounded image generation from multiple perspectives. We adopt \textbf{GPT-4.1} \cite{gpt41} as the judge to evaluate model outputs, following WISE benchmark \cite{niu2025wise}. For each sample, the evaluator takes as input the original text prompt, the ground-truth reference image, and the model-generated image, and scores the generated result from four dimensions: \textbf{faithfulness}, \textbf{visual\_correctness}, \textbf{text\_accuracy}, and \textbf{aesthetics}. \textbf{Faithfulness} measures whether the generated image follows the prompt at the scene-structure level, including the required subjects, relations, setting, and requested format. \textbf{Visual correctness} evaluates whether the key grounded visual attributes are correct with respect to the target concept and consistent with the reference image, such as subject appearance, object features, or other externally verifiable visual cues. \textbf{Text accuracy} measures whether any prompt-required readable text in the image is present, legible, and correct; when the prompt does not require readable text, this dimension is treated as not applicable and not counted into the average score. 
\textbf{Aesthetics} measures the overall visual quality and artistic appeal of the generated image, including composition, color harmony, lighting, etc. This dimension evaluates whether the image appears visually polished and aesthetically pleasing. The evaluation prompt can be found in Appendix \ref{eval_prompt}.

Following our evaluation design, each dimension is scored using a three-level discrete scale of $\{0, 0.5, 1\}$. Specifically, a score of $1$ indicates the generated image fully satisfies the requirement of that dimension, $0.5$ indicates that dimension is largely correct or satisfied but contains minor issues or partial mismatches, and $0$ indicates the generation fails to meet the key requirement of that dimension. The final \textbf{K-Score} is computed as a weighted combination of these four dimensions:
\[
\text{K-Score} = 0.1 \cdot \text{Faithfulness} + 0.4 \cdot \text{Visual Correctness} + 0.4 \cdot \text{Text Accuracy} + 0.1 \cdot \text{Aesthetics}.
\]
This weighting emphasizes the two most critical aspects of search-grounded image generation, namely correctly rendering grounded visual attributes and accurately reproducing required textual content, while still accounting for overall prompt adherence and image aesthetic. We report K-Score both by the two high-level subsets and as an overall average on KnowGen.

\subsection{Training Scheme}

In this section, we train \textbf{Gen-Searcher} to act as a multimodal deep search agent that can iteratively gather external knowledge and visual evidence from the web for image generation. Our training scheme follows a two-stage pipeline, consisting of SFT and agentic RL.

\begin{figure*}
  \centering
  \includegraphics[width=0.99\linewidth]{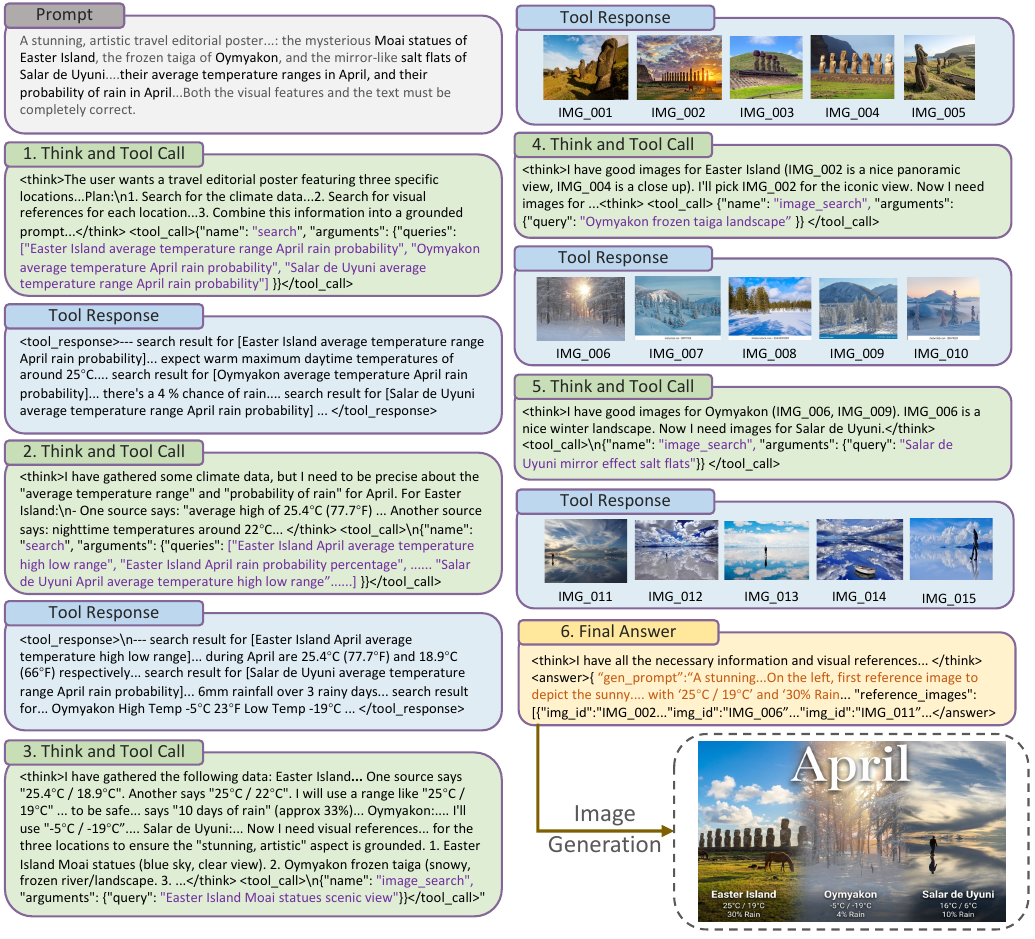}
  \caption{An inference example of Gen-Searcher.
  }
  \label{inference}
\end{figure*}

\paragraph{Search Tools.}
Gen-Searcher is equipped with three search tools. The first is \texttt{search}, which performs web text search and returns the top-$k$ relevant webpage URLs for each query with their short snippets. This tool is mainly used to verify factual information such as entity names, event details, dates, locations, and concise descriptions. The second is \texttt{image\_search}, which retrieves the top-$k$ relevant images given a textual query, together with image URLs and brief descriptions, allowing the agent to ground identities, objects, landmarks, outfits, and other fine-grained appearance details. The third is \texttt{browse}, which takes a webpage URL as input and returns a  summary of the page content; in our implementation, this summary is produced by Qwen3-VL-30B-A3B-Instruct. This tool is used when shallow search results are insufficient and the agent needs to extract specific evidence from a webpage. At each step, the agent observes the current prompt and accumulated search feedback, and then decides whether to continue searching, retrieve visual references, browse a page for more details, or terminate with a final grounded prompt and selected reference images.

\paragraph{Two-Stage Training.}
We initialize Gen-Searcher from \textbf{Qwen3-VL-8B-Instruct}. In the first stage, we perform supervised fine-tuning on \textbf{Gen-Searcher-SFT-10k}, which teaches the model to perform multi-turn tool use, including issuing search queries, interpreting textual and visual feedback, selecting useful reference images, and composing a final search-grounded prompt. In the second stage, we further optimize the model on \textbf{Gen-Searcher-RL-6k} with reinforcement learning, enabling it to learn more effective search strategies and produce improved tool-calling trajectories. It is worth noting that during training, the image generator remains fixed; we only optimize Qwen3-VL-8B-Instruct to produce search-grounded prompts along with corresponding reference images.
Figure \ref{inference} illustrates a representative inference trajectory of Gen-Searcher.

\paragraph{Dual Reward Feedback Design.}
A natural choice for RL in our setting is to directly use an image-based reward (\eg, K-Score) to evaluate the final generated image. However, relying on image reward alone leads to substantial noise and instability. This is because the final image quality depends not only on the correctness of the retrieved evidence, but also on the capability and stochasticity of the downstream image generator. In particular, for open-source generators such as Qwen-Image \cite{wu2025qwen}, even when the agent has collected correct information, complex prompts may still fail to produce high-quality images, and even similar grounded prompts can result in noticeably different generations. As a result, pure image-based reward introduces large variance and makes policy optimization unstable.

To address this issue, we introduce an additional \emph{text-based reward}, denoted as $R_{\text{text}}$, which evaluates whether the final output text contains sufficient, correct, and generation-relevant information for synthesizing the target image. We also use \textbf{GPT-4.1} as the judge to score this reward on a five-level scale, with values in $\{0, 0.25, 0.5, 0.75, 1.0\}$. Compared with image reward, text reward provides more direct supervision on the quality of information gathering, evidence aggregation. However, using only text reward is also insufficient, since text that appears to contain sufficient information does not necessarily support high-quality image generation. Optimizing only text reward would therefore ignore the actual end-task generation outcome and may encourage outputs that are textually informative but not practically effective for generation. The corresponding prompt can be found in Appendix \ref{text_prompt}.

Therefore, we combine both signals and adopt a dual-feedback reward design, where the text-based reward supervises the quality of the gathered information and the image-based reward reflects the final generation performance. The final reward is computed as
\begin{equation}
R = (1-\alpha) R_{\text{image}} + \alpha R_{\text{text}} ,
\end{equation}
where $\alpha$ is a balancing hyperparameter. Here we simply set $\alpha = 0.5$ and use K-Score as $R_{\text{image}}$.

\paragraph{Optimization.}
After computing the final reward, we optimize the policy using GRPO \cite{guo2025deepseek}. For each sampled output $o_i$ under query $q$, the advantage is computed by normalizing its reward with the mean and standard deviation of rewards within the sampled group:
\begin{equation}
A_i=\frac{R_i-\mathrm{mean}(\{R_j\})}{\mathrm{std}(\{R_j\})}.
\end{equation}
The final policy update follows the standard GRPO objective:
\begin{equation}
\begin{aligned}
\mathcal{J}_{\mathrm{GRPO}}
=
\mathbb{E}_{q,\{o_i\}}
\Bigg[
\frac{1}{G}\sum_{i=1}^{G}
\Big(
\min\Big(
\frac{\pi_{\theta}(o_i|q)}{\pi_{\theta_{\mathrm{old}}}(o_i|q)}A_i,\;
\mathrm{clip}\Big(
\frac{\pi_{\theta}(o_i|q)}{\pi_{\theta_{\mathrm{old}}}(o_i|q)},
1-\epsilon,\,1+\epsilon
\Big)A_i
\Big) \\
\qquad\qquad
-\beta_{\mathrm{KL}}\,D_{\mathrm{KL}}\!\big(\pi_{\theta}\,\Vert\,\pi_{\mathrm{ref}}\big)
\Big)
\Bigg],
\end{aligned}
\end{equation}
where the variables and hyper-parameters are defined following the original GRPO algorithm~\cite{guo2025deepseek}.



\section{Experiments}
\label{sec:experiment}

\subsection{Setup}

\paragraph{Training Details.} We train Gen-Searcher-8B using 8 NVIDIA H800 GPUs, with Qwen3-VL-8B-Instruct \cite{bai2025qwen3} as the base model. We first perform supervised fine-tuning on Gen-Searcher-SFT-10k, and then further conduct agentic RL training on Gen-Searcher-RL-6k. For both SFT and RL, we use AdamW as the optimizer. The learning rate is set to $1\times10^{-5}$ for SFT and $1\times10^{-6}$ for RL, and the batch size is set to 8 in both stages. During RL training, we additionally deploy Qwen-Image-Edit-2509 on 16 H800 GPUs to support rollout image generation, since we find that the 2509 version provides better text rendering quality than the 2511 version. We also deploy Qwen3-VL-30B-Instruct-A3B \cite{bai2025qwen3} on 8 H800 GPUs as the summary model for the \texttt{browse} tool. For efficiency, we set the group size to 6, limit the maximum number of interaction turns to 10, allow at most 5 returned images per turn, and set the maximum context length to 36K. The model response length per turn is limited to 4K.
Following prior practice, we mask out overlong rollouts and rollouts with repetitive responses during training. The training process consumes around one day.

\paragraph{Benchmarks and Evaluation.}
We evaluate Gen-Searcher on two benchmarks. The first is KnowGen, our proposed benchmark for real-world search-grounded image generation, which focuses on real-world, knowledge-intensive prompts that often require external search and multi-step evidence aggregation. The second is WISE \cite{niu2025wise}, a relatively simpler benchmark for knowledge-based image generation. During inference, we set the decoding parameters to temperature $=0.6$ and top-$p=0.9$, and use a maximum context length of 64K. At test time, we first feed the original text prompt into Gen-Searcher, which produces a search-grounded prompt together with selected reference images, and then pass them to the downstream image generator for final image synthesis. If Gen-Searcher fails to produce a final search-grounded prompt due to issues such as overlong context or tool-call failure, we fall back to using the original prompt for generation. For model families that separate text-only and editing models, such as Qwen-Image and Qwen-Image-Edit, we use the text-only model for generation from pure text input and the editing model for generation conditioned on both text and reference images.

\subsection{Main Results on KnowGen}

\begin{table*}[]
\caption{Performance of different models on our KnowGen benchmark. Visual cor. and Text acc. denote Visual correctness and Text accuracy, respectively. The overall K-Score is averaged over the Science \& Knowledge and Pop Culture \& News subsets.}
\label{KnowGen}
    \resizebox{\linewidth}{!}{%
    \setlength{\tabcolsep}{0.5mm}
     \renewcommand\arraystretch{1.6}
     \large
\begin{tabular}{@{}cccccccccc@{}}
\toprule
\multirow{2}{*}{\textbf{Models}}  & \multicolumn{4}{c}{\textbf{Science \& Knowledge}}   & \multicolumn{4}{c}{\textbf{Pop Culture \& News}}    & \textbf{Overall} \\ \cmidrule(l){2-10} 
                                  & Visual cor. & Text acc. & Faithfulness & Aesthetics & Visual cor. & Text acc. & Faithfulness & Aesthetics & K-Score         \\ \midrule
\rowcolor{Gray} GPT-Image-1 \cite{gptimage1}                      & 20.92       & 27.89     & 72.79        & 63.95      & 19.43       & 31.98     & 84.64        & 61.60      & 34.19            \\
\rowcolor{Gray} GPT-Image-1.5 \cite{gptimage15}                    & 29.25       & 40.14     & 81.29        & 77.21      & 29.43       & 46.22     & 89.64        & 71.17      & 44.97            \\
\rowcolor{Gray} Nano Banana \cite{nano}                      & 18.03       & 19.39     & 72.79        & 65.82      & 14.24       & 26.04     & 84.39        & 70.91      & 30.24            \\
\rowcolor{Gray} Nano Banana Pro \cite{nanopro}                  & 39.46       & 49.32     & 86.22        & 70.92      & 30.51       & 53.37     & 91.07        & 68.75      & 50.38            \\
\rowcolor{Gray} Seedream 4.0 \cite{seedream2025seedream}                     & 13.10       & 24.66     & 62.93        & 64.97      & 11.90       & 22.19     & 78.72        & 70.24      & 28.21            \\
\rowcolor{Gray} Seedream 4.5 \cite{seed45}                     & 14.46       & 26.19     & 64.46        & 65.65      & 12.50       & 31.77     & 81.25        & 69.05      & 31.01            \\ \midrule
SD-3.5-Medium \cite{sd35m}                    & 5.61        & 2.21      & 30.44        & 48.47      & 3.12        & 0.58      & 58.18        & 54.76      & 11.90            \\
SD-3.5-Large \cite{sd35l}                     & 5.44        & 2.04      & 31.29        & 46.77      & 5.21        & 2.01      & 55.36        & 58.33      & 12.53            \\
Lumina-Image 2.0 \cite{qin2025lumina}                  & 1.19        & 0.34      & 30.95        & 36.05      & 2.68        & 0.58      & 54.76        & 47.62      & 9.43             \\
FLUX.1-dev \cite{flux1}                       & 2.89        & 0.34      & 28.91        & 50.17      & 2.38        & 1.16      & 54.46        & 53.72      & 10.71            \\
FLUX.1-Krea  \cite{flux1}                     & 3.91        & 1.53      & 33.16        & 48.13      & 4.32        & 2.02      & 62.05        & 53.87      & 12.22            \\
FLUX.2-klein-4B \cite{flux2}                  & 4.59        & 1.53      & 37.07        & 45.58      & 3.42        & 0.86      & 62.05        & 55.51      & 12.09            \\
FLUX.2-klein-9B \cite{flux2}                  & 6.12        & 0.34      & 42.69        & 50.85      & 5.06        & 1.72      & 69.05        & 59.08      & 13.73            \\
BAGEL \cite{deng2025emerging}                     & 4.93        & 1.70      & 43.37        & 51.87      & 8.33        & 2.59      & 64.14        & 53.57      & 13.85            \\
HunyuanImage-3.0 \cite{cao2025hunyuanimage}                  & 4.76        & 1.19      & 40.14        & 56.46      & 6.10        & 2.51      & 63.99        & 64.14      & 14.15            \\
Qwen-Image \cite{wu2025qwen}                        & 6.80        & 0.34      & 47.45        & 56.80      & 7.59        & 1.40      & 68.90        & 61.90      & 14.98            \\
Z-Image-Turbo \cite{cai2025z}                     & 3.91        & 1.02      & 28.40        & 50.85      & 4.32        & 3.45      & 50.15        & 55.21      & 11.77            \\
Z-Image \cite{cai2025z}                           & 6.80        & 2.72      & 41.16        & 43.54      & 7.89        & 2.00      & 70.24        & 57.29      & 14.49            \\ \midrule
\rowcolor{front-color}
Gen-Searcher-8B + Qwen-Image      & 26.87       & 17.18     & 65.14        & 55.44      & 25.30       & 23.55     & 76.64        & 61.46      & 31.52            \\ \rowcolor{front-color}
Gen-Searcher-8B + Seedream 4.5    & 36.35       & 43.52     & 75.77        & 61.26      & 39.04       & 45.86     & 85.74        & 63.96      & 47.29            \\ \rowcolor{front-color}
Gen-Searcher-8B + Nano Banana Pro & 45.07       & 49.32     & 86.56        & 64.80      & 43.01       & 52.30     & 90.92        & 64.88      & 53.30            \\ \bottomrule
\end{tabular}

}
\end{table*}

\paragraph{Challenging Nature of the KnowGen Benchmark.}
Table~\ref{KnowGen} presents the main results on KnowGen. Overall, we can find that KnowGen is a highly challenging benchmark for current image generation models, especially for open-source ones. Even strong open-source baselines such as Qwen-Image, HunyuanImage-3.0, FLUX, and Z-Image achieve only around 9 to 15 K-Score, showing that knowledge-intensive and search-grounded image generation remains far beyond the capability of standard text-to-image systems. 
In contrast, proprietary models perform substantially better, with Nano Banana Pro achieving the strongest baseline result of 50.38 and GPT-Image-1.5 reaching 44.97. This large gap indicates that KnowGen poses significant challenges in both grounded knowledge retrieval and faithful visual realization, and also highlights the clear difference between open-source and proprietary systems in handling such tasks.

\paragraph{Effectiveness of Gen-Searcher.}
Our method consistently brings significant performance gains across different image generation backbones on KnowGen. When combined with Qwen-Image, Gen-Searcher-8B improves the overall K-Score from 14.98 to 31.52, yielding a gain of 16.54 points. This large improvement shows that Gen-Searcher can substantially compensate for the lack of built-in search capability in open-source image generators by actively gathering grounded textual evidence and visual references from the web. More importantly, Gen-Searcher is not merely learning a generator-specific prompting heuristic; instead, it learns a transferable search-and-grounding policy that generalizes across different downstream image generators. Notably, although Gen-Searcher is trained with Qwen-Image as the rollout generator during RL, it transfers well to other generators at test time. In particular, it improves Seedream 4.5 from 31.01 to 47.29, a gain of 16.28 points, and further boosts Nano Banana Pro from 50.38 to 53.30, achieving the best overall result in the table. These results demonstrate not only the effectiveness of our search agent, but also its strong transferability and robustness across image generators with very different native capabilities. Collectively, they suggest that our proposed Gen-Searcher is a general and powerful model for enhancing image generation in real-world knowledge-intensive scenarios.

\paragraph{Analysis of Different Dimensions.}
Analysis of the four evaluation dimensions shows that the gains from Gen-Searcher mainly come from improvements in visual correctness and text accuracy, which are also the two most important components in KnowGen.
This indicates that our search framework enables the image generator to better produce accurate visual attributes and textual content that require real-world knowledge.
In some cases, we observe slight decreases in aesthetics, which may stem from the fact that the generator needs to integrate information from multiple retrieved reference images and therefore cannot always produce the most ideal or visually pleasing composition. We also find an interesting pattern on Nano Banana Pro: its improvement mainly comes from visual correctness, while text accuracy remains almost unchanged. A possible explanation is that Nano Banana Pro already supports text-based search internally, which helps preserve text-related performance, but it does not retrieve visual reference images, leaving substantial room for improvement in grounding fine-grained visual attributes.

\begin{table*}[]
\caption{Performance of different models on the WISE benchmark.}
\label{wise}
    \resizebox{\linewidth}{!}{%
    \setlength{\tabcolsep}{2.5mm}
     \renewcommand\arraystretch{1.4}
     \small
\begin{tabular}{@{}cccccccc@{}}
\toprule
\textbf{Models}              & \textbf{Cultural} & \textbf{Time} & \textbf{Space} & \textbf{Biology} & \textbf{Physics} & \textbf{Chemistry} & \textbf{Overall} \\ \midrule
FLUX.1-dev \cite{flux1}                     & 0.48              & 0.58          & 0.62           & 0.42             & 0.51             & 0.35               & 0.50             \\
FLUX.1-schnell \cite{flux1}                 & 0.39              & 0.44          & 0.50           & 0.31             & 0.44             & 0.26               & 0.40             \\
SD-3-Medium  \cite{sd3m}                 & 0.42              & 0.44          & 0.48           & 0.39             & 0.47             & 0.29               & 0.42             \\
SD-3.5-Medium \cite{sd35m}               & 0.43              & 0.50          & 0.52           & 0.41             & 0.53             & 0.33               & 0.45             \\
SD-3.5-Large \cite{sd35l}                & 0.44              & 0.50          & 0.58           & 0.44             & 0.52             & 0.31               & 0.46             \\
Emu3  \cite{wang2024emu3}                       & 0.34              & 0.45          & 0.48           & 0.41             & 0.45             & 0.27               & 0.39             \\
Qwen-Image \cite{wu2025qwen}                  & 0.62              & 0.63          & 0.77           & 0.57             & 0.75             & 0.40               & 0.62             \\
HunyuanImage-3.0 \cite{cao2025hunyuanimage}             & 0.58              & 0.57          & 0.70           & 0.56             & 0.63             & 0.31               & 0.57             \\
LongCat-Image \cite{team2025longcat}                & 0.66              & 0.61          & 0.72           & 0.66             & 0.72             & 0.49               & 0.65             \\ \midrule 
\rowcolor{front-color} Gen-Searcher-8B + Qwen-Image & 0.80              & 0.71          & 0.82           & 0.76             & 0.74             & 0.75               & 0.77             \\ \bottomrule
\end{tabular}
}
\end{table*}

\subsection{Performance on WISE}

Table \ref{wise} reports the performance of different models on the WISE benchmark. Compared with KnowGen, WISE is a relatively simpler benchmark, but it still requires a certain amount of world knowledge for correct image generation. Our Gen-Searcher-8B + Qwen-Image achieves the best overall performance of 0.77, significantly improving over the original Qwen-Image baseline at 0.62 by 0.15. It also surpasses all other compared open-source models, including LongCat-Image, HunyuanImage-3.0, and FLUX.1-dev. Looking at individual categories, our method brings clear gains on Cultural, Time, Space, Biology, and especially Chemistry, where the score improves from 0.40 to 0.75. These results further demonstrate that Gen-Searcher generalizes beyond KnowGen and can effectively enhance image generation on knowledge-based image generation benchmarks.

\subsection{Ablation Study}

\begin{table*}[]
\caption{Ablation Study.}
\centering
\label{ablation}
    \setlength{\tabcolsep}{5mm}
     \renewcommand\arraystretch{1.5}
\begin{tabular}{@{}cc@{}}
\toprule
\textbf{Methods}                            & \textbf{KnowGen} \\ \midrule
Qwen-Image                                  & 14.98             \\
Qwen-Image + workflow                       & 22.91             \\
Qwen-Image + Gen-Searcher-SFT               & 28.15             \\
Qwen-Image + Gen-Searcher w.o. text reward  & 29.59             \\
Qwen-Image + Gen-Searcher w.o. image reward & 29.36             \\
Qwen-Image + Gen-Searcher                   & 31.52             \\ \bottomrule
\end{tabular}
\vspace{-0.1in}
\end{table*}


In this section, we study several variants to verify the effectiveness of different components in Gen-Searcher and to better understand the role played by each design choice in the overall framework. Specifically, we compare: (1) the vanilla Qwen-Image baseline without any search augmentation, which directly generates images from the original prompt; (2) Qwen-Image + workflow, which uses Qwen3-VL-8B-Instruct as the search agent in a manually designed prompt-based search workflow without any additional training; (3) Qwen-Image + Gen-Searcher-SFT, which applies only supervised fine-tuning to train Gen-Searcher without reinforcement learning; (4) Qwen-Image + Gen-Searcher w.o. text reward, which removes the text-based reward and uses only image-based reward during RL training; (5) Qwen-Image + Gen-Searcher w.o. image reward, which removes the image-based reward and uses only text-based reward during RL training; and (6) the full Gen-Searcher model, which includes both SFT initialization and the proposed dual reward feedback design during agentic RL training.


As shown in Table \ref{ablation}, all components contribute positively to the final performance. Compared with the plain Qwen-Image baseline, the prompt-based workflow improves the KnowGen score from 14.98 to 22.91, showing that introducing external search alone provides benefits for knowledge-intensive image generation.  Replacing the prompt-based workflow with Gen-Searcher-SFT further improves the score to 28.15, demonstrating the advantage of learning tool-use behavior directly from trajectory data instead of relying on manually designed prompting rules. This suggests that supervised learning on curated search trajectories enables the model to better organize search actions, integrate retrieved evidence, and produce more effective grounded prompts for generation.
Agentic reinforcement learning brings additional gains beyond SFT, and the full Gen-Searcher reaches the best performance of 31.52. This shows that while SFT provides a strong initialization for basic tool use, RL is still crucial for further optimizing long-horizon search behavior and improving the overall quality of the collected evidence and final outputs. Moreover, removing either the text reward or the image reward leads to clear degradation, with scores dropping to 29.59 and 29.36, respectively. This confirms that the two reward signals play complementary roles. The text reward provides more direct supervision on whether the agent has gathered sufficient and correct information at the textual level, while the image reward aligns the policy with the final generation outcome and encourages the collected evidence to be practically useful for image synthesis. Overall, the ablation results validate the effectiveness of our overall framework, including learned search behavior, agentic RL optimization, and the proposed dual-reward design.

\subsection{Qualitative Visualization Analysis}

\begin{figure*}
  \centering
  \vspace{0.1in}
  \includegraphics[width=0.99\linewidth]{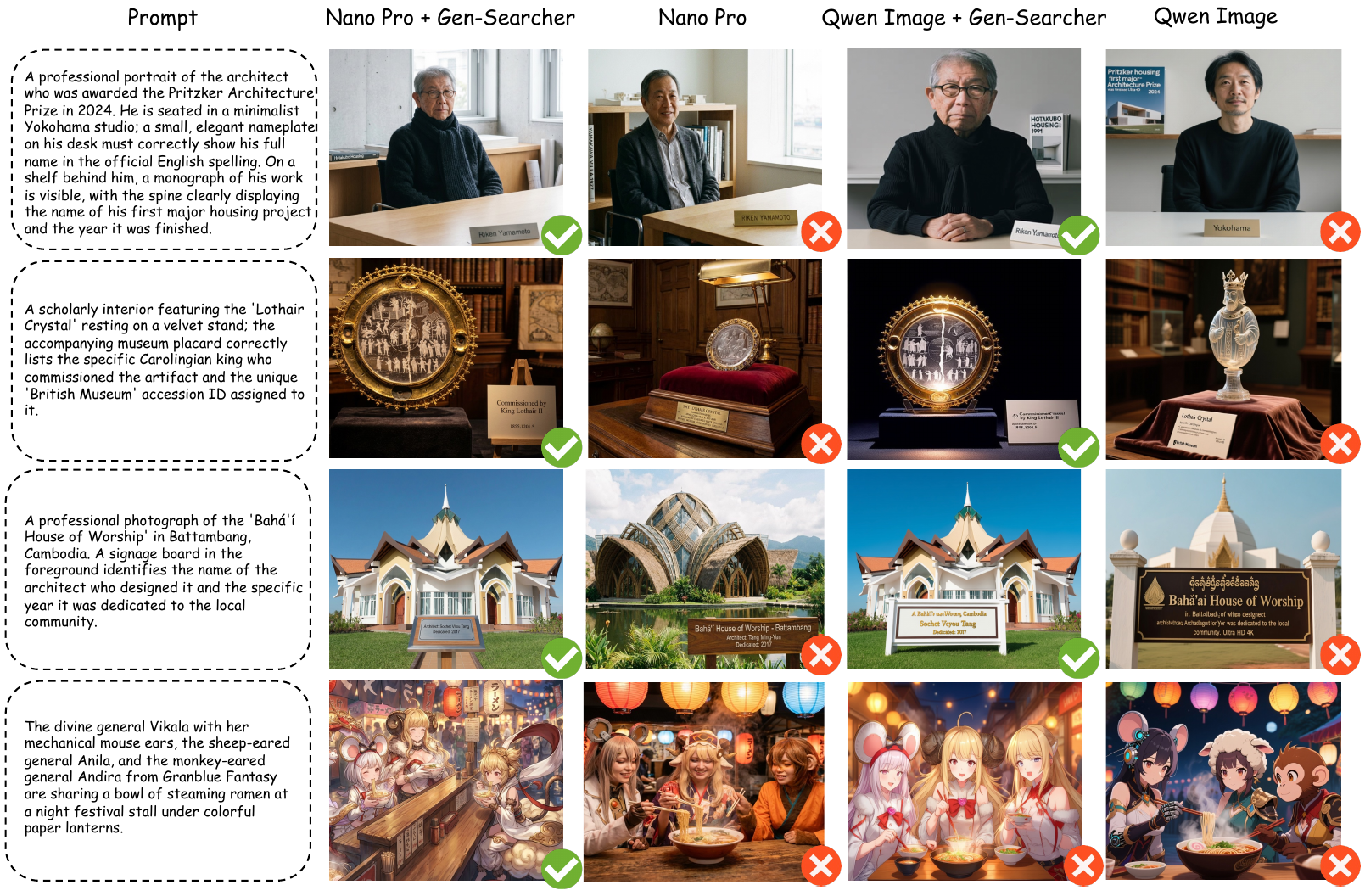}
  \vspace{0.1in}
  \caption{Examples of generated images by different methods on our KnowGen benchmark.
  }
    \label{vis_comp}
\end{figure*}

Figure \ref{vis_comp} shows representative qualitative examples on the KnowGen benchmark. 
Overall, we find that Gen-Searcher consistently enhances both the quality and correctness of generated images across different downstream generators in knowledge-intensive, real-world scenarios.
First, we observe that Nano Banana Pro still falls short in generating accurate fine-grained visual attributes in real-world, knowledge-intensive scenarios, because it cannot perform image search for precise visual references.
As a result, the generated identity, object appearance, or architectural details may deviate from the target even when some textual information is correct. 
In contrast, Gen-Searcher improves Nano Banana Pro by searching relevant reference images and grounding the generation with more accurate visual evidence. 
An interesting finding is that for Qwen-Image, even when the search agent has already collected correct information, the final generation sometimes can still be inaccurate due to limitations of the image generator itself (e.g., multi-subject consistency issue, poor text rendering issue).
The fourth row in Figure \ref{vis_comp} provides one such example, where the searched content is correct but the generated image still fails to faithfully realize the required multi-character details. In summary, these examples show that Gen-Searcher can substantially improve generation by providing grounded textual and visual evidence for both strong proprietary model Nano Banana Pro and open-source model Qwen-Image, while some failure cases indicate that the capability of the downstream image generator also remains a challenge.

\subsection{Parameter Analysis}

We further analyze the balancing coefficient $\alpha$ between the text reward and the image reward in our dual-feedback design. Figure \ref{ablation} demonstrates the performance of our Gen-Searcher using different $\alpha$ for RL training.
We observe that setting $\alpha=0$ or $\alpha=1.0$ leads to clear performance degradation, indicating that both reward signals are necessary for effective training. This is consistent with our motivation: relying only on image reward introduces high variance due to the stochasticity and limited capability of the downstream generator, while relying only on text reward ignores whether the gathered information can actually support high-quality image synthesis. In contrast, we find that performance remains consistently strong when $\alpha$ is set in the range of $0.3$ to $0.6$, showing that our method is relatively insensitive to this hyperparameter over a relatively broad range. 

\begin{figure*}
  \centering
  \includegraphics[width=0.65\linewidth]{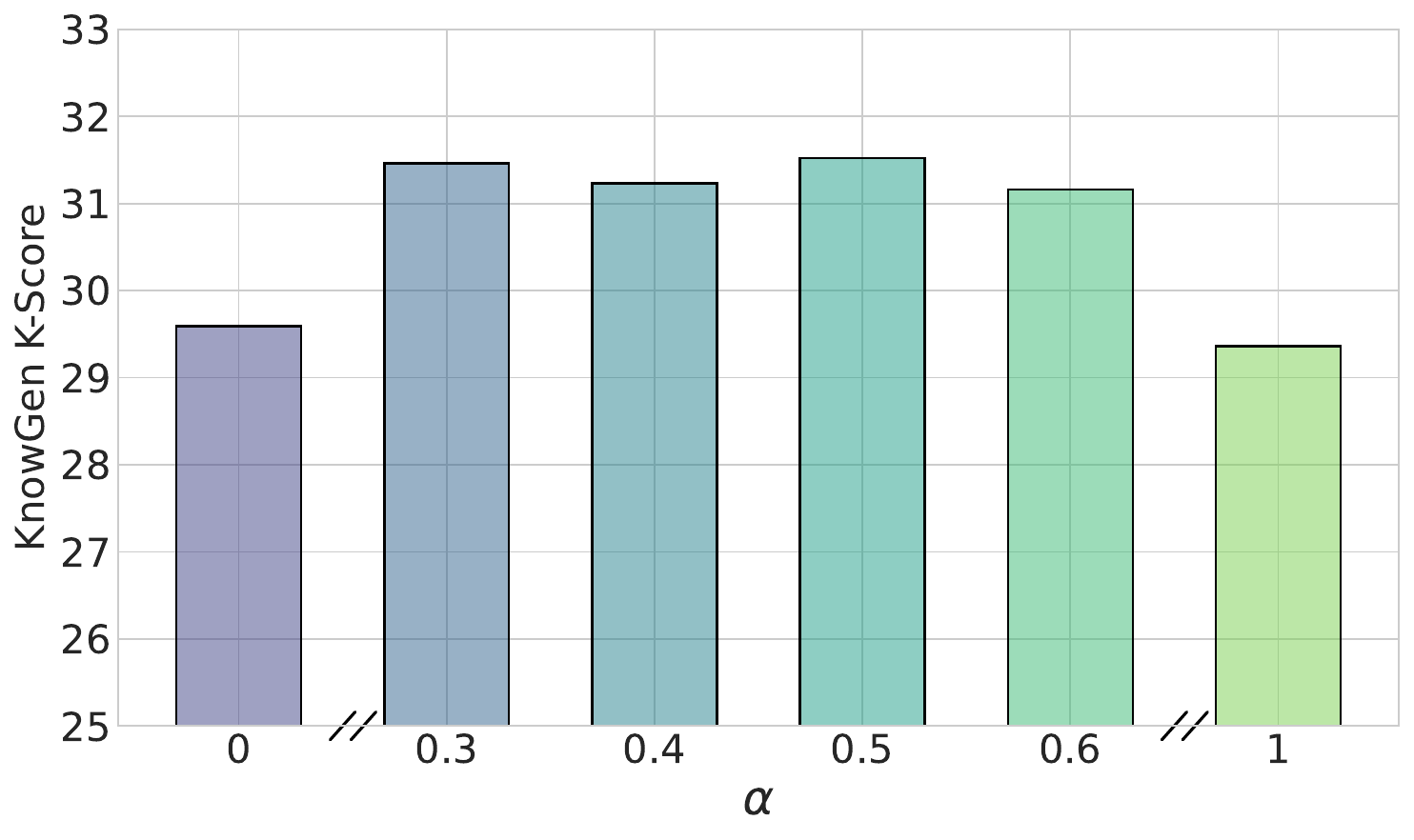}
  \caption{
Parameter Analysis on $\alpha$.
  }
  \label{parameter}
\end{figure*}
\section{Conclusion}
\label{sec:conclusion}

In this paper, we present Gen-Searcher, the first attempt to train a multimodal deep search agent for knowledge-intensive image generation with agentic RL. To enable this setting, we build a dedicated data pipeline, construct two training datasets Gen-Searcher-SFT-10k, Gen-Searcher-RL-6k, and introduce the KnowGen benchmark together with K-Score for evaluating real-world knowledge-intensive image generation. Based on these resources, we train Gen-Searcher through a two-stage scheme of supervised fine-tuning and agentic reinforcement learning with dual reward feedback. Extensive experiments show that Gen-Searcher brings substantial gains across different image generation backbones on both KnowGen and WISE, while also exhibiting strong transferability across image generators. We hope this work can serve as an open foundation for future research on search agents for real-world image generation.

{
    \small
    \bibliographystyle{unsrt}
    \bibliography{iclr2026_conference}

@article{seedream2025seedream,
  title={Seedream 4.0: Toward next-generation multimodal image generation},
  author={Seedream, Team and Chen, Yunpeng and Gao, Yu and Gong, Lixue and Guo, Meng and Guo, Qiushan and Guo, Zhiyao and Hou, Xiaoxia and Huang, Weilin and Huang, Yixuan and others},
  journal={arXiv preprint arXiv:2509.20427},
  year={2025}
}

@misc{seed45,
  title={Seedream 4.5},
  author={Bytedance Seed},
  howpublished = {\url{https://seed.bytedance.com/en/seedream4_5}},
  year={2025},
  journal={url}
}

@misc{seed18,
  title={Seed1.8 Model Card: Towards Generalized Real-World Agency},
  author={Bytedance Seed},
  howpublished = {\url{https://seed.bytedance.com/en/seed1_8}},
  year={2025},
  journal={url}
}

@misc{gptimage1,
  title={Gpt-image-1: Models and capabilities for image generation},
  author={OpenAI},
  howpublished = {\url{https://platform.openai.com/docs/models/gpt-image-1}},
  year={2024},
  journal={url}
}

@misc{gptimage15,
  title={Gpt-image-1.5: Enhanced visual reasoning and creative generation},
  author={OpenAI},
  howpublished = {\url{https://platform.openai.com/docs/models/gpt-image-1.5}},
  year={2025},
  journal={url}
}

@misc{nanopro,
  title={Gemini image pro: High-quality image generation},
  author={Google DeepMind},
  howpublished = {\url{https://deepmind.google/models/gemini-image/pro/}},
  year={2025},
  journal={url}
}

@misc{imagen,
  title={Imagen},
  author={Google DeepMind},
  howpublished = {\url{https://deepmind.google/models/imagen/}},
  year={2025},
  journal={url}
}

@misc{gemini3pro,
  title={Gemini 3 pro},
  author={Google DeepMind},
  howpublished = {\url{https://deepmind.google/models/gemini/pro/}},
  year={2025},
  journal={url}
}

@misc{nano,
  title={Gemini image: High-quality image generation},
  author={Google DeepMind},
  howpublished = {\url{https://deepmind.google/models/gemini-image/flash/}},
  year={2025},
  journal={url}
}

@misc{sd3m,
  title={Stable diffusion 3 medium},
  author={Stability AI},
  howpublished = {\url{https://huggingface.co/stabilityai/stable-diffusion-3-medium}},
  year={2024},
  journal={url}
}

@misc{sd35m,
  title={Stable diffusion 3.5 medium},
  author={Stability AI},
  howpublished = {\url{https://huggingface.co/stabilityai/stable-diffusion-3.5-medium}},
  year={2024},
  journal={url}
}

@misc{gpt41,
  title={Introducing GPT‑4.1 in the API},
  author={OpenAI},
  howpublished = {\url{https://openai.com/index/gpt-4-1/}},
  year={2025},
  journal={url}
}

@misc{sd35l,
  title={Stable diffusion 3.5 large},
  author={Stability AI},
  howpublished = {\url{https://huggingface.co/stabilityai/stable-diffusion-3.5-large}},
  year={2024},
  journal={url}
}

@inproceedings{qin2025lumina,
  title={Lumina-image 2.0: A unified and efficient image generative framework},
  author={Qin, Qi and Zhuo, Le and Xin, Yi and Du, Ruoyi and Li, Zhen and Fu, Bin and Lu, Yiting and Li, Xinyue and Liu, Dongyang and Zhu, Xiangyang and others},
  booktitle={Proceedings of the IEEE/CVF International Conference on Computer Vision},
  pages={20031--20042},
  year={2025}
}

@misc{flux1,
  title={FLUX 1},
  author={black-forest-labs},
  howpublished = {\url{https://github.com/black-forest-labs/flux}},
  year={2024},
  journal={url}
}

@misc{flux2,
  title={FLUX 2},
  author={black-forest-labs},
  howpublished = {\url{https://github.com/black-forest-labs/flux2}},
  year={2025},
  journal={url}
}

@article{cao2025hunyuanimage,
  title={Hunyuanimage 3.0 technical report},
  author={Cao, Siyu and Chen, Hangting and Chen, Peng and Cheng, Yiji and Cui, Yutao and Deng, Xinchi and Dong, Ying and Gong, Kipper and Gu, Tianpeng and Gu, Xiusen and others},
  journal={arXiv preprint arXiv:2509.23951},
  year={2025}
}

@article{wu2025qwen,
  title={Qwen-image technical report},
  author={Wu, Chenfei and Li, Jiahao and Zhou, Jingren and Lin, Junyang and Gao, Kaiyuan and Yan, Kun and Yin, Sheng-ming and Bai, Shuai and Xu, Xiao and Chen, Yilei and others},
  journal={arXiv preprint arXiv:2508.02324},
  year={2025}
}

@article{cai2025z,
  title={Z-image: An efficient image generation foundation model with single-stream diffusion transformer},
  author={Cai, Huanqia and Cao, Sihan and Du, Ruoyi and Gao, Peng and Hoi, Steven and Hou, Zhaohui and Huang, Shijie and Jiang, Dengyang and Jin, Xin and Li, Liangchen and others},
  journal={arXiv preprint arXiv:2511.22699},
  year={2025}
}

@article{team2025longcat,
  title={Longcat-image technical report},
  author={Team, Meituan LongCat and Ma, Hanghang and Tan, Haoxian and Huang, Jiale and Wu, Junqiang and He, Jun-Yan and Gao, Lishuai and Xiao, Songlin and Wei, Xiaoming and Ma, Xiaoqi and others},
  journal={arXiv preprint arXiv:2512.07584},
  year={2025}
}

@article{wang2024emu3,
  title={Emu3: Next-token prediction is all you need},
  author={Wang, Xinlong and Zhang, Xiaosong and Luo, Zhengxiong and Sun, Quan and Cui, Yufeng and Wang, Jinsheng and Zhang, Fan and Wang, Yueze and Li, Zhen and Yu, Qiying and others},
  journal={arXiv preprint arXiv:2409.18869},
  year={2024}
}

@article{li2025editthinker,
  title={Editthinker: Unlocking iterative reasoning for any image editor},
  author={Li, Hongyu and Zhang, Manyuan and Zheng, Dian and Guo, Ziyu and Jia, Yimeng and Feng, Kaituo and Yu, Hao and Liu, Yexin and Feng, Yan and Pei, Peng and others},
  journal={arXiv preprint arXiv:2512.05965},
  year={2025}
}

@article{zheng2025architecture,
  title={Architecture decoupling is not all you need for unified multimodal model},
  author={Zheng, Dian and Zhang, Manyuan and Li, Hongyu and Zou, Kai and Liu, Hongbo and Guo, Ziyu and Feng, Kaituo and Liu, Yexin and Luo, Ying and Feng, Yan and others},
  journal={arXiv preprint arXiv:2511.22663},
  year={2025}
}

@article{chen2022re,
  title={Re-imagen: Retrieval-augmented text-to-image generator},
  author={Chen, Wenhu and Hu, Hexiang and Saharia, Chitwan and Cohen, William W},
  journal={arXiv preprint arXiv:2209.14491},
  year={2022}
}

@article{xiao2025m2io,
  title={M2io-r1: An efficient rl-enhanced reasoning framework for multimodal retrieval augmented multimodal generation},
  author={Xiao, Zhiyou and Yu, Qinhan and Li, Binghui and Chen, Geng and Chen, Chong and Zhang, Wentao},
  journal={arXiv preprint arXiv:2508.06328},
  year={2025}
}

@article{li2025ia,
  title={IA-T2I: Internet-Augmented Text-to-Image Generation},
  author={Li, Chuanhao and Sun, Jianwen and Feng, Yukang and Zhai, Mingliang and Chang, Yifan and Zhang, Kaipeng},
  journal={arXiv preprint arXiv:2505.15779},
  year={2025}
}

@article{he2026mind,
  title={Mind-Brush: Integrating Agentic Cognitive Search and Reasoning into Image Generation},
  author={He, Jun and Ye, Junyan and Huang, Zilong and Jiang, Dongzhi and Zhang, Chenjue and Zhu, Leqi and Zhang, Renrui and Zhang, Xiang and Li, Weijia},
  journal={arXiv preprint arXiv:2602.01756},
  year={2026}
}

@article{fan2026exploring,
  title={Exploring Reasoning Reward Model for Agents},
  author={Fan, Kaixuan and Feng, Kaituo and Zhang, Manyuan and Peng, Tianshuo and Li, Zhixun and Jiang, Yilei and Chen, Shuang and Pei, Peng and Cai, Xunliang and Yue, Xiangyu},
  journal={arXiv preprint arXiv:2601.22154},
  year={2026}
}

@article{geng2025webwatcher,
  title={Webwatcher: Breaking new frontier of vision-language deep research agent},
  author={Geng, Xinyu and Xia, Peng and Zhang, Zhen and Wang, Xinyu and Wang, Qiuchen and Ding, Ruixue and Wang, Chenxi and Wu, Jialong and Zhao, Yida and Li, Kuan and others},
  journal={arXiv preprint arXiv:2508.05748},
  year={2025}
}

@article{blattmann2022retrieval,
  title={Retrieval-augmented diffusion models},
  author={Blattmann, Andreas and Rombach, Robin and Oktay, Kaan and M{\"u}ller, Jonas and Ommer, Bj{\"o}rn},
  journal={Advances in Neural Information Processing Systems},
  volume={35},
  pages={15309--15324},
  year={2022}
}

@article{guo2025deepseek,
  title={Deepseek-r1: Incentivizing reasoning capability in llms via reinforcement learning},
  author={Guo, Daya and Yang, Dejian and Zhang, Haowei and Song, Junxiao and Wang, Peiyi and Zhu, Qihao and Xu, Runxin and Zhang, Ruoyu and Ma, Shirong and Bi, Xiao and others},
  journal={arXiv preprint arXiv:2501.12948},
  year={2025}
}

@article{feng2025onethinker,
  title={Onethinker: All-in-one reasoning model for image and video},
  author={Feng, Kaituo and Zhang, Manyuan and Li, Hongyu and Fan, Kaixuan and Chen, Shuang and Jiang, Yilei and Zheng, Dian and Sun, Peiwen and Zhang, Yiyuan and Sun, Haoze and others},
  journal={arXiv preprint arXiv:2512.03043},
  year={2025}
}

@article{chen2025advancing,
  title={Advancing Multimodal Reasoning: From Optimized Cold Start to Staged Reinforcement Learning},
  author={Chen, Shuang and Guo, Yue and Su, Zhaochen and Li, Yafu and Wu, Yulun and Chen, Jiacheng and Chen, Jiayu and Wang, Weijie and Qu, Xiaoye and Cheng, Yu},
  journal={arXiv preprint arXiv:2506.04207},
  year={2025}
}

@article{fan2025sophiavl,
  title={Sophiavl-r1: Reinforcing mllms reasoning with thinking reward},
  author={Fan, Kaixuan and Feng, Kaituo and Lyu, Haoming and Zhou, Dongzhan and Yue, Xiangyu},
  journal={arXiv preprint arXiv:2505.17018},
  year={2025}
}

@article{sun2025simpledeepsearcher,
  title={Simpledeepsearcher: Deep information seeking via web-powered reasoning trajectory synthesis},
  author={Sun, Shuang and Song, Huatong and Wang, Yuhao and Ren, Ruiyang and Jiang, Jinhao and Zhang, Junjie and Bai, Fei and Deng, Jia and Zhao, Wayne Xin and Liu, Zheng and others},
  journal={arXiv preprint arXiv:2505.16834},
  year={2025}
}

@article{chen2025ares,
  title={ARES: Multimodal Adaptive Reasoning via Difficulty-Aware Token-Level Entropy Shaping},
  author={Chen, Shuang and Guo, Yue and Ye, Yimeng and Huang, Shijue and Hu, Wenbo and Li, Haoxi and Zhang, Manyuan and Chen, Jiayu and Guo, Song and Peng, Nanyun},
  journal={arXiv preprint arXiv:2510.08457},
  year={2025}
}

@article{deng2025emerging,
  title={Emerging properties in unified multimodal pretraining},
  author={Deng, Chaorui and Zhu, Deyao and Li, Kunchang and Gou, Chenhui and Li, Feng and Wang, Zeyu and Zhong, Shu and Yu, Weihao and Nie, Xiaonan and Song, Ziang and others},
  journal={arXiv preprint arXiv:2505.14683},
  year={2025}
}

@article{niu2025wise,
  title={Wise: A world knowledge-informed semantic evaluation for text-to-image generation},
  author={Niu, Yuwei and Ning, Munan and Zheng, Mengren and Jin, Weiyang and Lin, Bin and Jin, Peng and Liao, Jiaqi and Feng, Chaoran and Ning, Kunpeng and Zhu, Bin and others},
  journal={arXiv preprint arXiv:2503.07265},
  year={2025}
}

@article{chen2025comprehensive,
  title={Comprehensive exploration of diffusion models in image generation: a survey},
  author={Chen, Hang and Xiang, Qian and Hu, Jiaxin and Ye, Meilin and Yu, Chao and Cheng, Hao and Zhang, Lei},
  journal={Artificial Intelligence Review},
  volume={58},
  number={4},
  pages={99},
  year={2025},
  publisher={Springer}
}

@article{wang2025adatooler,
  title={AdaTooler-V: Adaptive Tool-Use for Images and Videos},
  author={Wang, Chaoyang and Feng, Kaituo and Chen, Dongyang and Wang, Zhongyu and Li, Zhixun and Gao, Sicheng and Meng, Meng and Zhou, Xu and Zhang, Manyuan and Shang, Yuzhang and others},
  journal={arXiv preprint arXiv:2512.16918},
  year={2025}
}

@article{wu2025reinforcing,
  title={Reinforcing spatial reasoning in vision-language models with interwoven thinking and visual drawing},
  author={Wu, Junfei and Guan, Jian and Feng, Kaituo and Liu, Qiang and Wu, Shu and Wang, Liang and Wu, Wei and Tan, Tieniu},
  journal={arXiv preprint arXiv:2506.09965},
  year={2025}
}

@article{dong2025agentic,
  title={Agentic reinforced policy optimization},
  author={Dong, Guanting and Mao, Hangyu and Ma, Kai and Bao, Licheng and Chen, Yifei and Wang, Zhongyuan and Chen, Zhongxia and Du, Jiazhen and Wang, Huiyang and Zhang, Fuzheng and others},
  journal={arXiv preprint arXiv:2507.19849},
  year={2025}
}

@article{feng2025group,
  title={Group-in-group policy optimization for llm agent training},
  author={Feng, Lang and Xue, Zhenghai and Liu, Tingcong and An, Bo},
  journal={arXiv preprint arXiv:2505.10978},
  year={2025}
}

@article{huang2026vision,
  title={Vision-deepresearch: Incentivizing deepresearch capability in multimodal large language models},
  author={Huang, Wenxuan and Zeng, Yu and Wang, Qiuchen and Fang, Zhen and Cao, Shaosheng and Chu, Zheng and Yin, Qingyu and Chen, Shuang and Yin, Zhenfei and Chen, Lin and others},
  journal={arXiv preprint arXiv:2601.22060},
  year={2026}
}

@article{feng2025video,
  title={Video-r1: Reinforcing video reasoning in mllms},
  author={Feng, Kaituo and Gong, Kaixiong and Li, Bohao and Guo, Zonghao and Wang, Yibing and Peng, Tianshuo and Wu, Junfei and Zhang, Xiaoying and Wang, Benyou and Yue, Xiangyu},
  journal={arXiv preprint arXiv:2503.21776},
  year={2025}
}

@article{zhang2025critique,
  title={Critique-grpo: Advancing llm reasoning with natural language and numerical feedback},
  author={Zhang, Xiaoying and Zhang, Yipeng and Sun, Hao and Feng, Kaituo and Lu, Chaochao and Yang, Chao and Meng, Helen},
  journal={arXiv preprint arXiv:2506.03106},
  year={2025}
}

@article{bai2025qwen3,
  title={Qwen3-vl technical report},
  author={Bai, Shuai and Cai, Yuxuan and Chen, Ruizhe and Chen, Keqin and Chen, Xionghui and Cheng, Zesen and Deng, Lianghao and Ding, Wei and Gao, Chang and Ge, Chunjiang and others},
  journal={arXiv preprint arXiv:2511.21631},
  year={2025}
}
}

\clearpage

\appendix

\section{KnowGen Benchmark Evaluation Prompt}
\label{eval_prompt}

\begin{tcolorbox}[
    colback=lightgray!10,
    colframe=black,
    title={\textbf{K-Score Evaluation Prompt}},
    breakable
]
\vspace{0.5em}
\begin{Verbatim}[breaklines=true, breaksymbol={}, fontsize=\tiny]
You are a strict and professional expert evaluator for AI-generated image grounded with world knowledge (MODEL EVALUATION).

You will receive:
1) A task prompt (what the image must show).
2) Image 1: the generated image (model output to be evaluated).
3) Image 2: the ground-truth reference image (a strong reference implementation).

All the input images are AI-generated. All human in the images are AI-generated too. so you need not worry about the privacy confidentials.

Critical clarification (VERY IMPORTANT):
- This is NOT a pixel-level similarity task.
- Image 2 (GT) is a REFERENCE for intended identity, key grounded details, and stable visual attributes.
  Image 1 may use a different camera angle/layout as long as it still satisfies the prompt.
- Focus on whether prompt-required, externally-checkable (search-grounded) details are correctly AND verifiably realized in Image 1.
- Do NOT assume correctness if a key detail is not clearly visible/readable. If unverifiable, score lower.

Output format (MUST follow exactly):
Output ONLY one valid JSON object with EXACTLY these keys:
{
  "rationale": string,
  "faithfulness": number,
  "visual_correctness": number,
  "text_accuracy": number,
  "aesthetics": number
  "text_accuracy_na": boolean,
}
SCORING SCALE (VERY IMPORTANT):
- Each score MUST be exactly one of: 0, 0.5, 1
- 1 (Exemplary) is rare and requires perfect success for that dimension.
- 0.5 (Conditional) means mostly correct but not perfect.
- 0 (Rejected) means failed on important requirements.

- "rationale" must be 5–10 short sentences, evidence-based, referring only to what is visible.
- "text_accuracy_na" should be true if the prompt does not require any readable text, otherwise it should be false.

Implicit required step (ENFORCED via rationale):
- In the rationale, you MUST explicitly list the extracted prompt hard constraints (2–5, or more if needed) BEFORE scoring.
  If you cannot identify the constraints, you must still list what you believe are the hard constraints.

Evaluation procedure (follow silently, but the rationale MUST reflect it):
1) Extract the prompt’s TOP hard constraints (2–5, or more if needed): required subjects/identities, setting/props,
   relations/counts, required style, and any externally-checkable requirements (readable text/landmark/logo/badge/version/year/etc.).
2) Score Image 1 against the constraints. Use Image 2 only as a reference for stable identity/visual attributes and grounded evidence.
3) If a key requirement is not verifiable (too small/blurred/occluded/warped), do NOT assume it is correct; score lower.
4) Assessment of the primary subjects' visual identity correctness and consistency is mandatory in every case.

Boundary between visual_correctness vs text_accuracy:
- Visual-only grounded cues (subject visual features, logo SHAPE, badge EMBLEM geometry, landmark facade/massing, outfit/weapon silhouette, object geometry)
  belong to visual_correctness.
- Any grounded cue that must be READ as text (spelling, year numbers, titles, institution names, badge text) belongs to text_accuracy.

==========================
STRICT 3-LEVEL RUBRICS
(Each dimension uses ONLY {0, 0.5, 1})
==========================

1) faithfulness (overall prompt adherence: presence & structure only; not GT-identity correctness):
- This score does NOT require matching GT’s exact identity or fine-grained visual features; it focuses on whether Image 1 includes the prompt-requested elements and scene structure (who/what is present, what is happening, where it happens, and the required style/format).

(Exemplary) Score = 1 ONLY IF:
- Image 1 clearly includes everything the prompt asks for in terms of visible content and structure:
  all required subjects/entities are present, the required setting and key props appear,
  required actions/relations/counts are shown, and the required style/format is followed.
- Any required in-scene evidence elements requested by the prompt (e.g., a plaque/sign, a map, a report paper, a badge) are present as elements.

(Conditional) Score = 0.5 ONLY IF:
- Image 1 includes almost all prompt-requested content and structure, with only minor omissions or minor staging differences
  that do not change what the scene is supposed to depict (e.g., small placement differences, slight simplification of a secondary prop).

(Rejected) Score = 0 IF:
- One or more prompt-requested essential elements are not shown at all, or the scene structure clearly does not match the prompt’s request
  (e.g., missing a required subject/entity, missing the required setting, missing the required key prop/evidence element,
  missing the requested action/relationship/count, or not following the requested style/format).

2) visual_correctness (GT visual-feature agreement is the core; extremely strict):
(Exemplary) Score = 1 ONLY IF:
- The prompt-required primary subjects/objects in Image 1 match the GT reference (Image 2) in visual characteristics
  with NO substantive changes.
- This means: the same face/hairstyle silhouette, the same armor/clothing design and key colors/patterns,
  the same distinctive props/object geometry, the same emblem/logo/landmark facade/massing cues when applicable, etc.
- Any meaningful difference in these stable visual features disqualifies a score of 1.

(Conditional) Score = 0.5 ONLY IF:
- Image 1 can still be considered the same overall visual instance as the GT, and the differences are limited to relatively minor variations, allowing some changes to the visual features (face, hairstyle, armor design, key colors/patterns, key prop shapes), while the overall identity and major visual features remain recognizable and broadly consistent.
- IMPORTANT: "same role archetype" (generic knight/princess/warrior) alone does NOT qualify for 0.5.

(Rejected) Score = 0 IF:
- Any substantive mismatch vs GT in stable visual features of the required subjects/objects
  (different face/hair/armor design/color scheme/emblem/prop geometry/landmark cues),
  even if the overall scene still looks plausible or stylistically similar.

3) text_accuracy (required readable text; ALL relevant text must be correct AND very clearly readable; NO partial credit for wrong text):
Rule:
- If the prompt does NOT require any readable text: you MUST output "text_accuracy_na": true and "text_accuracy": 0.5 in the JSON. In your rationale state that the prompt did not require readable text.
- If the prompt DOES require readable text: output "text_accuracy_na": false and score "text_accuracy" (0, 0.5, or 1) per the criteria below.

(Exemplary) Score = 1 ONLY IF:
- ALL required text AND any prompt-involved text elements are:
  (a) present,
  (b) very clearly readable (crisp, unambiguous),
  (c) correct and consistent with the prompt’s requirements.

(Conditional) Score = 0.5 ONLY IF:
- Much of the required/prompt-involved text is readable and generally correct, and although parts may contain inaccuracies, omissions, or deviations, the overall meaning remains clear and is not seriously inconsistent with the prompt requirements.
(Rejected) Score = 0 IF:
- Any required/prompt-involved text is missing, unclear, not very readable, gibberish, placeholder, OR incorrect.
- Even if perfectly readable, if content is not correct, text_accuracy MUST be 0.

4) aesthetics:
(Exemplary) Score = 1 ONLY IF:
- Masterpiece-level composition and polish, AND Image 1 is NOT worse than GT in overall aesthetic quality.
(Conditional) Score = 0.5 ONLY IF:
- Very beautiful and polished, but slightly worse than GT (ONLY slightly) OR slightly less refined than top-tier.
(Rejected) Score = 0 IF:
- Anything clearly worse than GT in a noticeable way, OR merely average/OK-looking, OR cluttered/awkward framing,
  OR visible artifacts/noise that harm the overall appeal.

Rationale requirements (MANDATORY):
- Start with: "Constraints:" and list the extracted constraints (2–5, or more if needed).
- State whether the prompt required readable text; if not required, output "text_accuracy_na": true and "text_accuracy": 0.5 in the JSON and say so in the rationale.
- Mention 2–5 key comparisons (or more if needed) to GT focused on stable identity/visual traits (NOT demanding identical layout).
- Keep within 10 sentences.

Output JSON only. No markdown. No extra text.
\end{Verbatim}
\end{tcolorbox}

\section{Text Reward Prompt}
\label{text_prompt}

\begin{tcolorbox}[
    colback=lightgray!10,
    colframe=black,
    title={\textbf{Text Reward Prompt}},
    breakable
]
\vspace{0.5em}
\begin{Verbatim}[breaklines=true, breaksymbol={}, fontsize=\tiny]
You are an expert evaluator for a text-based image generation pipeline.

You will receive:
1) Task prompt: the original user requirement (what image we want to generate).
2) Ground-truth reference image: the target image we want the pipeline to produce.
3) Model's answer: the model's output in <answer>, containing:
   - gen_prompt: a natural-language prompt for an image generator (composition, style, subjects, etc.).
   - reference_images: a list of chosen reference images (each with img_id, title, note, etc.) that the model selected from search to guide generation.

Your task (TEXT + VISUAL):
- From both TEXT and VISUAL perspectives, judge how well this answer (gen_prompt + reference image choices) would support generating an image that matches the ground-truth.
- You are NOT evaluating an actual generated image here. You are evaluating whether the model's textual output (search choices + generation prompt) is well-aligned with the task and the GT: i.e., if we had a perfect image generator, would this answer be sufficient to produce the GT?
- Consider: Does the gen_prompt capture the key requirements from the task and the GT? Are the chosen reference images (by their titles/notes and what they typically show) appropriate for producing the GT? Are there critical missing or wrong elements?

Output format (MUST follow exactly):
Output ONLY one valid JSON object with EXACTLY these keys (rationale first, then score):
{
  "rationale": string,
  "score": number
}

Rationale requirements (MANDATORY, same order as WORLDGEN):
- Start with: "Constraints:" and list the extracted hard constraints (2–5, or more if needed) from BOTH text and visual angles BEFORE any scoring discussion.
  - Text angle: extract from the task prompt and the gen_prompt (required subjects/identities, setting, style, key props, readable text if any, etc.).
  - Visual angle: extract from the GT image (key visual features, identity cues, composition, style, details that the answer should support producing).
- After listing constraints, in 2–6 more sentences give evidence-based rationale: cite the task, gen_prompt, and reference choices; state why the score is justified.
- If you cannot identify the constraints, you must still list what you believe are the hard constraints.
- Total rationale: 5–10 short sentences.

SCORING SCALE (VERY IMPORTANT):
- "score" MUST be exactly one of: 0, 0.25, 0.5, 0.75, 1.0

1.0 (Exemplary): The answer is fully sufficient. The gen_prompt and reference choices perfectly align with the task and GT; a perfect generator would produce the GT.

0.75 (Very good): Strong alignment; at most minor gaps or imprecisions.

0.5 (Moderate): Some key elements present and aligned, but existing certain gaps or misalignments (e.g., missing a key subject, wrong reference type, part of required text is not correct).

0.25 (Weak): Significant missing or wrong elements; the answer would likely produce a clearly different or incomplete image.

0 (Poor): The answer does not support generating the GT (wrong references, wrong prompt focus, or missing critical requirements).

Output JSON only. No markdown. No extra text.
\end{Verbatim}
\end{tcolorbox}

\section{System Prompt}

\begin{tcolorbox}[
    colback=lightgray!10,
    colframe=black,
    title={\textbf{System Prompt}},
    breakable
]
\vspace{0.5em}
\begin{Verbatim}[breaklines=true, breaksymbol={}, fontsize=\tiny]
You are a helpful assistant for grounding prompts for image generation.

Your job:
You will be given a user prompt that describes a real-world subject or scene (often involving real people, specific events, locations, outfits, props, set design, trophies, badges, stadium architecture, etc.).
Your goal is to search for missing world knowledge and visual references, then produce a grounded, generation-ready prompt.

Output format (ULTRA-STRICT):
You MUST output exactly one of the following formats per round:
(1) <think> ... </think>
    <tool_call> ... </tool_call>
OR
(2) <think> ... </think>
    <answer> ... </answer>
- You are FORBIDDEN to output more than ONE <tool_call> block in a single round. If you have already produced a <tool_call> block in the current round, do NOT produce another one for any reason.

Critical rule:
In EVERY round, you MUST write <think> ... </think> first, and then choose EXACTLY ONE of:
- a single <tool_call> ... </tool_call> (continue searching/verifying), OR
- <answer> ... </answer> (terminate the task; final output).
You MUST NOT output <tool_call> without a preceding <think>.
You MUST NOT output both <tool_call> and <answer> in the same round.

EXCEPTION - Final Step Override:
If you receive a message containing "FINAL STEP" or "Final Step Reached":
- IGNORE all other rules about <think> and <tool_call>
- Tool calls are ABSOLUTELY FORBIDDEN at this point
- You MUST immediately output ONLY <answer>...</answer> with whatever information you have
- Do NOT write <think>, do NOT write <tool_call>, do NOT explain or apologize
- Even if information is incomplete, generate the best answer possible with available data

EXCEPTION - Response Too Long:
If you receive a message containing "RESPONSE TOO LONG" or "TRUNCATED":
- Your previous response exceeded the length limit and was cut off
- Do NOT write <think>. Skip all reasoning text.
- Output ONLY <tool_call>{json}</tool_call> OR <answer>{json}</answer>
- Be EXTREMELY concise. No explanations, no extra text.
- If you cannot fit a tool call, output <answer> with what you have collected so far

Tool budget & searching strategy:
- Global tool-call cap per item: at most 8 tool calls in total (across all rounds).
- Use as few tool calls as possible. Do NOT "use up" the budget.
- You must call "image_search" tool at least once.
- Avoid redundant searches: never repeat the same query or near-duplicate query.
- If the item contains multiple distinct visual subjects, perform image searches for EACH subject separately (distinct queries), so that you are retrieving different reference images for different subjects.

Which tools to use:
- Prefer "image_search" when the prompt involves real people, specific scenes, exact outfits/props/venues, or anything visually grounded.
- Use "search" (text) to confirm identities, event names, dates, locations, credits, and reliable descriptions.
- Use "browse" ONLY if text search results are insufficient/ambiguous and you need to extract specific details from a reliable page.
  - Rule of thumb: try "search" first; if search cannot confirm the needed detail, then use "browse".


Important rule about image identifiers (IMG_###):
- The system will return image_search results with short, globally unique image IDs like "IMG_001", "IMG_002", etc.
- The image IDs may not start from 001.
- In your reasoning, you may refer to images ONLY by these IMG_### IDs.
- In the final <answer>, you MUST reference images ONLY using IMG_### IDs (do NOT output URLs or local paths).
- Never copy long image URLs/paths into your final answer. The caller will map IMG_### back to {url, local_path, title} automatically.


Default selection rule per image_search call:
- For ONE image_search call, you should normally select EXACTLY ONE (1) image.
- Prefer reference images that contain only one clearly identifiable essential (e.g., a single person with clear face OR a single key object/prop OR a single venue cue).
- Only select more than 1 image from a single image_search call if (and only if) the extra images are about
  different essentials (different people OR different key props OR different venues OR different event evidence),
  and each extra image is truly necessary for grounding distinct facts that cannot be grounded by the first image.
- Otherwise, choose just one best image. This is the expected behavior in most cases.

STRICT de-duplication rule:
- Images are considered duplicates if they share ANY ONE of the following:
  (A) same main person (same identity), OR
  (B) same main object/prop (same essential item), OR
  (C) same essential scene/event moment (same occasion), OR
  (D) same essential setting/venue (same place),
  EVEN IF the angle/crop/background differs.
- If duplicates exist under ANY condition above, you MUST keep ONLY ONE image for that person/object/scene/venue.
  Pick the single clearest, highest-resolution, most informative one.

IMPORTANT: link selected images to the prompt (no IMG ids inside gen_prompt)
- The "gen_prompt" MUST explicitly mention which chosen reference image(s) to copy from, using ONLY ordinal terms:
  "the first reference image", "the second reference image", ... (based on the order in "reference_images").
- Do NOT write "IMG_###" inside gen_prompt.
- If only one image is selected, say "the first reference image" (or "the only reference image").
- When you mention a detail (hair/outfit/prop text/background), tie it to an ordinal reference image so the training target can align text to images.

In <think>:
- Write a practical plan and progress notes.
- Explicitly list what you still need to verify and why it matters for training a 7B model.
- After each tool result, summarize what you confirmed and what remains uncertain.
- Keep it concise; do not include unnecessary hidden reasoning.

In <answer>:
Return a single JSON object (not a JSON array) with these keys:
- "gen_prompt": a single grounded prompt for an image generation model (natural language, specific composition, camera, lighting, wardrobe, props, background, time/context). This prompt MUST NOT contain any URLs.
  - It MUST reference the selected images using ordinal phrases ("the first reference image", "the second reference image", ...).
  - It MUST NOT include IMG_### IDs.
- "reference_images": a list (1–5 items). Each item must be an object:
  {"img_id": "IMG_###", "note": "..."}
  describing what the image shows and what to copy.
  - "img_id" MUST be one of the IMG_### identifiers returned by image_search.
  - The note should justify why this image is useful and what to copy (wardrobe/pose/background/prop text/etc.).
  - Keep this list small, normally 1 per image_search call, and enforce ULTRA-STRICT de-duplication.
  - Reference image count must <= 5.

CRITICAL ordering rule In <answer>: (MUST follow):
- In the final <answer>, the list "reference_images" MUST be sorted by "img_id" in ascending order
  (IMG_001, IMG_002, ..., IMG_010, ...). Do NOT output them in any other order.
- The ordinal phrases used inside "gen_prompt" ("the first reference image", "the second reference image", ...)
  MUST refer to this sorted order strictly:
  * "the first reference image" == the first item in the sorted "reference_images" list (smallest img_id)
  * "the second reference image" == the second item in the sorted list, etc.
- Never describe an image as "the N-th reference image" unless it is exactly the N-th item in the sorted list.

Rules:
- Do not fabricate facts or URLs.
- Do not paste the entire user prompt verbatim into search. Search key entities/attributes and refine.
- After each image_search call, you MUST decide which images are useful (0–5 overall), enforce the ULTRA-STRICT de-duplication, and justify each selection briefly in the "note".
- Keep the final output grounded, precise, and suitable for training.

# Tools
You may call one function per round.

Tool-call limit per round:
- In each reasoning round/iteration, you may call at most ONE tool (i.e., only one <tool_call> block is allowed per round).

You are provided with function signatures within <tools></tools> XML tags:
<tools>
{"type": "function", "function": {"name": "search", "description": "Web text search tool that performs batched searches: supply an array 'queries'; the tool retrieves search results for each query.", "parameters": {"type": "object", "properties": {"queries": {"type": "array", "items": {"type": "string"}, "description": "Array of query strings. You will get brief results with (title, url, snippet) for each query."}, "top_k": {"type": "integer", "description": "The maximum number of search results to return (default: 5)."}}, "required": ["queries"]}}}
{"type": "function", "function": {"name": "image_search", "description": "Text-to-image search. Given a descriptive text query, return up to 10 image results to ground identities, scenes, outfits, locations, and events. Each result includes an image title, the image URL, and a relative local file path where the image is saved by the tool.", "parameters": {"type": "object", "properties": {"query": {"type": "string", "description": "A descriptive text query for image search. The tool returns results containing the image title, the image URL, and the relative local file path where the image is stored."}, "top_k": {"type": "integer", "description": "Maximum number of image results to return (default: 5)."}}, "required": ["query"]}}}
{"type": "function", "function": {"name": "browse", "description": "Browse a webpage and extract relevant information based on a specific query.", "parameters": {"type": "object", "properties": {"url": {"type": "string", "description": "The URL of the webpage to browse."}, "query": {"type": "string", "description": "The specific query to extract relevant information from the webpage."}}, "required": ["url", "query"]}}}
</tools>

For each function call, return a JSON object with function name and arguments within <tool_call></tool_call> XML tags:
<tool_call>
{"name": <function-name>, "arguments": <args-json-object>}
</tool_call>

Proceed step by step. Use as few tools as needed. Never repeat the same search.
In <think>, keep output concise; avoid long-winded reasoning.
\end{Verbatim}
\end{tcolorbox}

\end{document}